%% file: main.tex
\documentclass[10pt,twocolumn,letterpaper]{article}

\usepackage[dvipsnames, wave]{xcolor}
\usepackage{pgfplots}
\usepackage{pgfplotstable}
\usepackage{pgffor}

\usepackage{iccv}
\usepackage{times}
\usepackage{epsfig}
\usepackage{graphicx}
\usepackage{amsmath}
\usepackage{amssymb}
\usepackage{soul}

\graphicspath{ {figs/} }

\usepackage{amssymb}
\usepackage{booktabs}
\usepackage{makecell}
\usepackage{tabu,multirow}
\usepackage{pbox}
\usepackage{multicol}
\usepackage{hhline}
\usepackage{subcaption}
\usepackage{enumitem}
\usepackage[normalem]{ulem}
\usepackage{xspace}

\usepackage{wasysym}

\input{tex/pgf_color.tex}

\newcommand{\tbf}[1]{{\textbf{#1}}}

\newcommand\hmm[1]{\ifnum\ifhmode\spacefactor\else2000\fi>1000 \uppercase{#1}\else#1\fi}

\newcommand{\hiConvPrefix}[0]{Oct\xspace}

\newcommand{\hiConv}[0]{OctConv\xspace}

\newcommand{\hiConvName}[0]{Octave Convolution\xspace}

\newcommand{\hierarchicalfeaturerepresentation}[0]{\hmm{m}ulti-frequency feature representation\xspace}

\newcommand{\myparagraph}[1]{\vspace{1pt}\noindent\textbf{#1.}~}

\newcommand{\hk}[0]{k}
\newcommand{\wk}[0]{k}

\usepackage{xr}
\iccvfinalcopy 

\def\iccvPaperID{**} 

\ificcvfinal\fi
\begin{document}

\title{Drop an Octave: Reducing Spatial Redundancy in \\ Convolutional Neural Networks with \hiConvName}

\author{
  Yunpeng Chen$^\dagger$$^\ddagger$, Haoqi Fan$^\dagger$, Bing Xu$^\dagger$, Zhicheng Yan$^\dagger$, Yannis Kalantidis$^\dagger$, \\
  Marcus Rohrbach$^\dagger$, Shuicheng Yan$^\ddagger$$^\flat$, Jiashi Feng$^\ddagger$ \\
  $^\dagger$Facebook AI, $^\ddagger$National University of Singapore, $^\flat$Yitu Technology \\
}


\maketitle

\input{tex/s0_abstract.tex}

\input{tex/s1_intro.tex}


\input{tex/s2_related.tex}

\input{tex/s3_method.tex}


\input{tex/s4_implement.tex}

\input{tex/s5_exp.tex}


\input{tex/s7_conclusion.tex}

{\small
\bibliographystyle{ieee_fullname}
\bibliography{main}
}

\newpage
\setcounter{section}{0}
\setcounter{subsection}{0}

\input{supp/content.tex}


\end{document}

%% file: tex/pgf_color.tex
\colorlet{MyColor}{black}%
\newcommand{\MixValue}{0}
\newcommand*{\SetColor}[1]{%
    \ifnum#1<21
        \pgfmathtruncatemacro{\MixValue}{100*#1/20}%
        \colorlet{MyColor}{orange!\MixValue!red}%
    \else
        \ifnum#1<41
            \pgfmathtruncatemacro{\MixValue}{100*(#1-20)/20}%
            \colorlet{MyColor}{yellow!\MixValue!orange}%
        \else
            \ifnum#1<61
                \pgfmathtruncatemacro{\MixValue}{100*(#1-40)/20}%
                \colorlet{MyColor}{green!\MixValue!yellow}%
            \else
                \ifnum#1<81
                    \pgfmathtruncatemacro{\MixValue}{100*(#1-60)/20}%
                    \colorlet{MyColor}{blue!\MixValue!green}%
                \else
                    \ifnum#1<101
                        \pgfmathtruncatemacro{\MixValue}{100*(#1-80)/20}%
                        \colorlet{MyColor}{violet!\MixValue!blue}%
                    \else
                    \fi%
                \fi%
           \fi%
       \fi%
    \fi%
}%

%% file: tex/s0_abstract.tex
\begin{abstract}
In natural images, information is conveyed at different frequencies where higher frequencies are usually encoded with fine details and lower frequencies are usually encoded with global structures. Similarly, the output feature maps of a convolution layer can also be seen as a mixture of information at different frequencies. 
In this work, we propose to factorize the mixed feature maps by their frequencies, and design a novel \hiConvName (\hiConv) operation\footnote{\url{https://github.com/facebookresearch/OctConv}} to store and process feature maps that vary spatially ``slower'' at a lower spatial resolution reducing both memory and computation cost. Unlike existing multi-scale methods, \hiConv is formulated as a single, generic, plug-and-play convolutional unit that can be used as a direct replacement of (vanilla) convolutions without any adjustments in the network architecture. It is also orthogonal and complementary to methods that suggest better topologies or reduce channel-wise redundancy like group or depth-wise convolutions.
We experimentally show that by simply replacing convolutions with \hiConv, we can consistently boost accuracy for both image and video recognition tasks, while reducing memory and computational cost. An \hiConv-equipped ResNet-152 can achieve 82.9\% top-1 classification accuracy on ImageNet with merely 22.2 GFLOPs.
\end{abstract}

%% file: tex/s1_intro.tex
\section{Introduction}

The efficiency of Convolutional Neural Networks (CNNs) keeps increasing with recent efforts to reduce the inherent redundancy in dense model parameters~\cite{DSD, ThiNet,ClipQ} and in the channel dimension of feature maps~\cite{ResNeXt,MobileNetV1,multifiber,xception}. However, substantial redundancy also exists in the spatial dimension of the feature maps produced by CNNs, where each location stores its own feature descriptor independently, while ignoring common information between adjacent locations that could be stored and processed together.

\begin{figure}[t!]
    \begin{subfigure}[b]{0.95\columnwidth}
      \includegraphics[width=\textwidth]{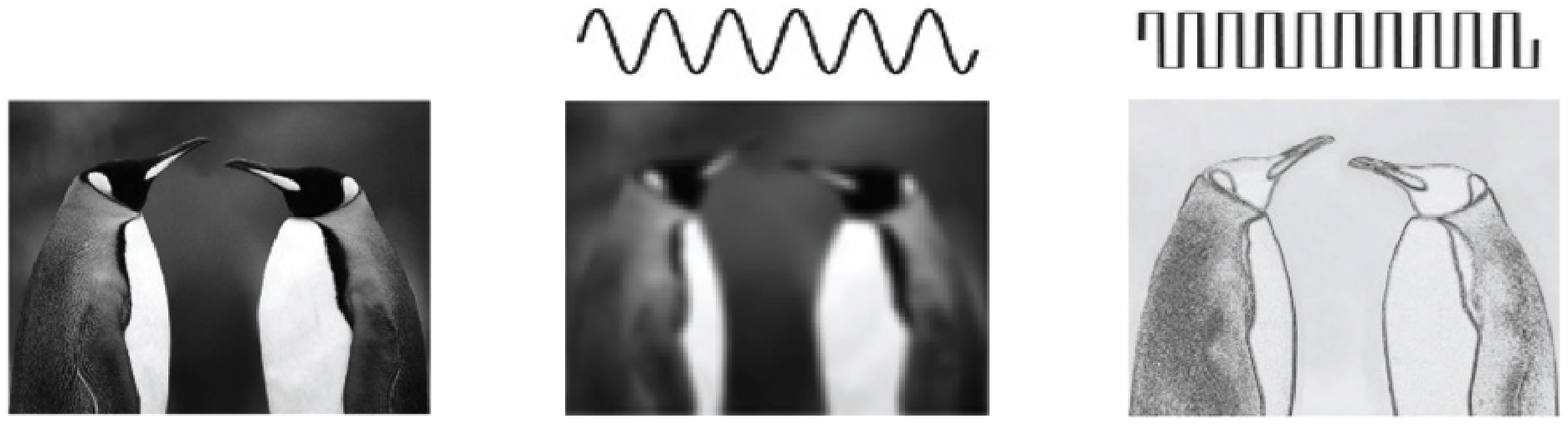}
      \caption{Separating the low and high spatial frequency signal~\cite{campbell1968application, spatialvision}.}
      \label{fig:teaser_a}
    \end{subfigure}
    \begin{subfigure}[b]{.3\columnwidth}
      \includegraphics[width=\textwidth]{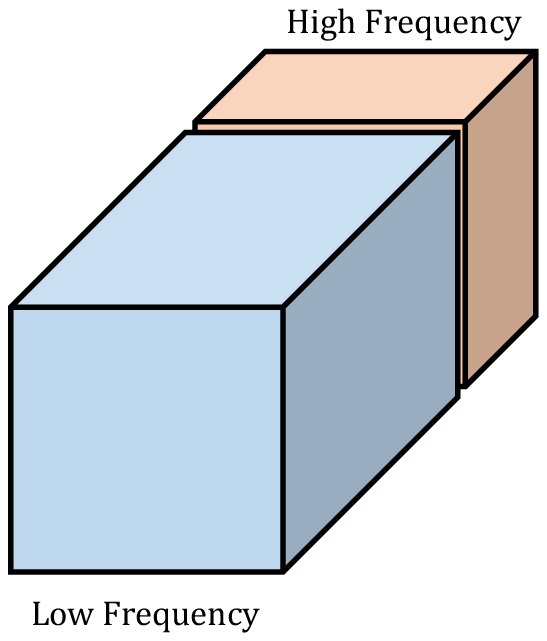}
      \caption{}
    \end{subfigure}
    \begin{subfigure}[b]{.3\columnwidth}
      \includegraphics[width=\textwidth]{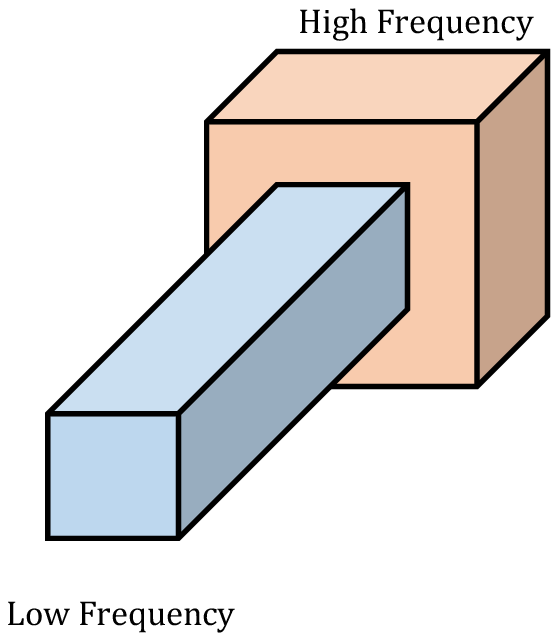}
      \caption{}
    \end{subfigure}
    \begin{subfigure}[b]{.34\columnwidth}
      \includegraphics[width=.9\textwidth]{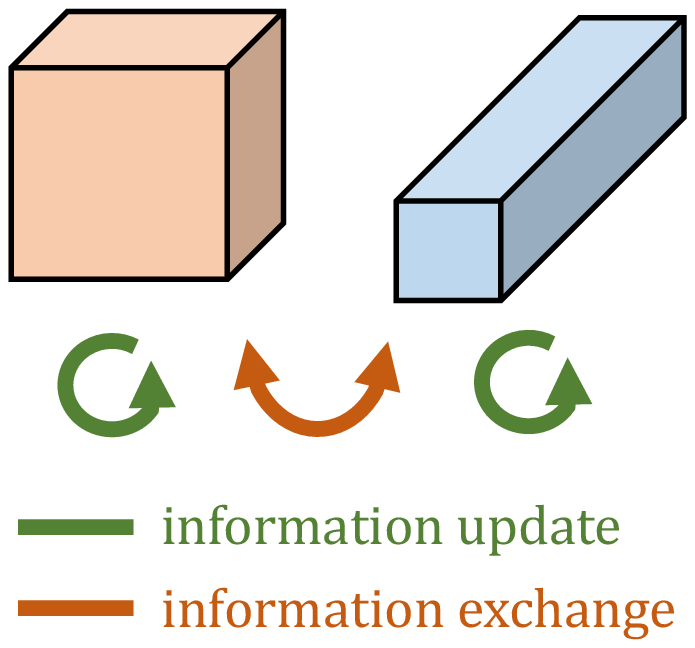}
      \caption{}
    \end{subfigure}
    \vspace{-2mm}
    \caption{(a)~Motivation. The spatial frequency model for vision~\cite{campbell1968application, spatialvision} shows that natural image can be decomposed into a low and a high spatial frequency part. (b)~The output maps of a convolution layer can also be factorized and grouped by their spatial frequency. (c) The proposed multi-frequency feature representation stores the smoothly changing, low-frequency maps in a low-resolution tensor to reduce spatial redundancy. (d) The proposed \hiConvName operates directly on this representation. It updates the information for each group and further enables information exchange between groups.}
    \label{fig:motivation}
    \vspace{-5mm}
  \end{figure}

As shown in Figure~\ref{fig:motivation}(a), a natural image can be decomposed into a low spatial frequency component that describes the smoothly changing structure and a high spatial frequency component that describes the rapidly changing fine details~\cite{campbell1968application, spatialvision,stephane1999wavelet,sweldens1998lifting}. Similarly, we argue that the output feature maps of a convolution layer can also be decomposed into features of different spatial frequencies and propose a novel \hierarchicalfeaturerepresentation which stores high- and low-frequency feature maps into different groups as shown in Figure~\ref{fig:motivation}(b). Thus, the spatial resolution of the low-frequency group can be safely reduced by sharing information between neighboring locations to reduce spatial redundancy as shown in Figure~\ref{fig:motivation}(c). To accommodate the novel feature representation, we generalize the vanilla convolution, and propose \emph{\hiConvName (\hiConv)} which takes in feature maps containing tensors of two frequencies one octave apart, and extracts information directly from the low-frequency maps without the need of decoding it back to the high-frequency as shown in Figure~\ref{fig:motivation}(d). As a replacement of vanilla convolution, \hiConv consumes substantially less memory and computational resources. In addition, \hiConv processes low-frequency information with corresponding (low-frequency) convolutions and effectively enlarges the receptive field in the original pixel space and thus can  improve recognition performance.


We design the \hiConv in a generic way, making it a plug-and-play replacement for the vanilla convolution. 
Since \hiConv mainly focuses on processing feature maps at multiple spatial frequencies and reducing their spatial redundancy, it is orthogonal and complementary to existing methods that focus on building better CNN topology~\cite{densenet, InceptionV1, vgg, AmoebaNet, PNASNet}, reducing channel-wise redundancy in convolutional feature maps~\cite{ResNeXt,xception,MobileNetV2,ShuffleNetV2, CondenseNet} and reducing redundancy in dense model parameters~\cite{ClipQ, DSD, ThiNet}. Moreover, different from methods that exploit multi-scale information~\cite{bLNet,ELASTIC,SlowFast}, \hiConv can be easily deployed as a plug-and-play unit to replace convolution, without the need of changing network architectures or requiring hyper-parameters tuning. Compared to the closely related Multi-grid convolution~\cite{ke2017multigrid}, \hiConv provides more insights on reducing the spatial redundancy in CNNs based on the frequency model and adopts more efficient inter-frequency information exchange strategy with better performance. We further integrate the \hiConv into a wide variety of backbone architectures (including the ones featuring group, depth-wise, and 3D convolutions) and demonstrate universality of \hiConv.

Our experiments demonstrate that by simply replacing the vanilla convolution with \hiConv, we can consistently improve the performance of popular 2D CNN backbones including ResNet~\cite{ResNetV1,ResNetV2}, ResNeXt~\cite{ResNeXt}, DenseNet~\cite{densenet}, MobileNet~\cite{MobileNetV1,MobileNetV2} and SE-Net~\cite{SENet} on 2D image recognition on ImageNet~\cite{imagenet}, as well as 3D CNN backbones C2D~\cite{nonlocal} and I3D~\cite{nonlocal} on video action recognition on Kinetics~\cite{kay2017kinetics, k400, k600}. The \hiConv-equipped \hiConvPrefix-ResNet-152 can match or outperform state-of-the-art manually designed networks~\cite{ShuffleNetV2,SENet} at lower memory and computational cost.

\noindent
Our contributions can be summarized as follows:
\begin{itemize}[leftmargin=*,noitemsep,nolistsep]
\item  We propose to factorize convolutional feature maps into two groups at different spatial frequencies and process them with different convolutions at their corresponding frequency, one octave apart. As the resolution for low frequency maps can be reduced, this saves both storage and computation. This also helps each layer gain a larger receptive field to capture more contextual information.
\item  We design a plug-and-play operation named \hiConv to replace the vanilla convolution for operating on the new feature representation directly and reducing spatial redundancy. Importantly, \hiConv is fast in practice and 
achieves a speedup close to the theoretical limit.
\item  We extensively study the properties of the proposed \hiConv on a variety of backbone CNNs for image and video tasks and achieve significant performance gain even comparable to the best AutoML networks.
\end{itemize}

%% file: tex/s2_related.tex
\section{Related Work}

\myparagraph{Improving the efficiency of CNNs}
Ever since the pioneering work on AlexNet~\cite{alexnet} and VGG~\cite{vgg}, researchers have made substantial efforts to improve the efficiency of CNNs. ResNet~\cite{ResNetV1,ResNetV2} and DenseNet~\cite{densenet} improve the network topology by adding shortcut connections to early layers. ResNeXt~\cite{ResNeXt} and ShuffleNet~\cite{ShuffleNetV1} use sparsely connected group convolutions to reduce redundancy in inter-channel connectivity. Xception~\cite{xception} and MobileNet~\cite{MobileNetV1,MobileNetV2} adopt depth-wise convolutions that further reduce the connection density. Meanwhile, NAS~\cite{NASNet}, PNAS~\cite{PNASNet} and AmoebaNet~\cite{AmoebaNet} propose to atomically find the best network topology for a given task. Pruning methods, such as DSD~\cite{DSD} and ThiNet~\cite{ThiNet}, focus on reducing the redundancy in the model parameters by eliminating the least significant weight or connections in CNNs. Besides, HetConv~\cite{HetConv} propose to replace the vanilla convolution filters with heterogeneous convolution filters that are in different sizes. However, all of these methods ignore the redundancy on the spatial dimension of feature maps, which is addressed by the proposed \hiConv, making \hiConv orthogonal and complementary to these previous methods. Noticeably, \hiConv does not change the connectivity between feature maps, making it also different from inception-alike multi-path designs~\cite{InceptionV1,InceptionV4,ResNeXt}.

\myparagraph{Multi-scale Representation Learning}
Prior to the success of deep learning, multi-scale representation has long been applied for local feature extraction, such as the SIFT features~\cite{lowe2004distinctive}. In the deep learning era, multi-scale representation also plays a important role due to its strong robustness and generalization ability. FPN~\cite{FPN} and PSP~\cite{PSP} merge convolutional features from different depths at the end of the networks for object detection and segmentation tasks. MSDNet~\cite{huang2018multi} and HR-Nets~\cite{sun2019deep}, proposed carefully designed network architectures that contain multiple branches where each branch has it own spatial resolution. The bL-Net~\cite{bLNet} and ELASTIC-Net~\cite{ELASTIC} adopt similar idea, but are designed as a replacement of residual block for ResNet~\cite{ResNetV1,ResNetV2} and thus are more flexible and easier to use. But extra expertise and hyper-parameter tuning are still required when adopt them to architectures beyond ResNet, such as MobileNetV1~\cite{MobileNetV1}, DenseNet~\cite{densenet}. 
Multi-grid CNNs~\cite{ke2017multigrid} propose a multi-grid pyramid feature representation and define the MG-Conv operator as a replacement of convolution operator, which is conceptually similar to our method but is motivated for exploiting multi-scale features. Compared with MG-Conv, \hiConv adopts more efficient design to exchange inter-frequency information with higher performance as can be found in Sec.~\ref{sec:imp:mg-conv} and Sec.~\ref{sec:exp:imgcls:sota}. 
For video models, the recently proposed SlowFast Networks~\cite{SlowFast} introduce multi-scale pathways on the temporal dimension. As we show in Section~\ref{sec:exp:video}, this is complementary to \hiConv which operates on the spatial dimensions.

In a nutshell, \hiConv focuses on reducing the spatial redundancy in CNNs and is designed to replace vanilla convolution operations without needing to adjust backbone CNN architecture. We extensively compare \hiConv to closely related methods in the sections of method and experiment and show that \hiConv CNNs give top results on a number of challenging benchmarks.

%% file: tex/s3_method.tex
\section{Method}

In this section, we first introduce the octave feature representation and then describe  \hiConvName, which operates directly on it. We also discuss implementation details and show how to integrate \hiConv into group and depth-wise convolution architectures.

\subsection{Octave Feature Representation}
\label{sec:feature_representation}

For the vanilla convolution, all input and output feature maps have the same spatial resolution, which may not be necessary since some of the feature maps may represent low-frequency information which is spatially redundant and can be further compressed as illustrated in Figure~\ref{fig:motivation}. 

To reduce the spatial redundancy, we introduce the \emph{octave feature representation} that explicitly factorizes the feature map tensors into groups corresponding to low and high frequencies. The scale-space theory~\cite{lindeberg2013scale} provides us with a principled way of creating scale-spaces of spatial resolutions, and defines an \emph{octave} as a division of the spatial dimensions by a power of $2$ (we only explore $2^1$ in this work). We follow this fashion and reduce the spatial resolution of the low-frequency feature maps by an octave.

Formally, let $X \in \mathbb{R}^{c\times h \times w}$ denote the input feature tensor of a convolutional layer, where $h$ and $w$ denote the spatial dimensions and $c$ the number of feature maps or channels. We explicitly factorize $X$ along the channel dimension into $X = \{X^H,X^L\}$, where the high-frequency feature maps $X^H \in \mathbb{R}^{(1-\alpha)c \times  h \times  w}$ capture fine details and the low-frequency maps $X^L \in \mathbb{R}^{\alpha c\times \frac{h}{2} \times \frac{w}{2}}$  vary slower in the spatial dimensions (w.r.t. the image locations). Here $\alpha \in [0,1]$ denotes the \emph{ratio} of channels allocated to the low-frequency part and the low-frequency feature maps are defined \textit{an octave lower} than the high frequency ones, \ie at half of the spatial resolution as shown in Figure~\ref{fig:motivation}(c).

In the next subsection, we introduce a convolution operator that operates directly on this \hierarchicalfeaturerepresentation and name it \emph{\hiConvName} (\emph{\hiConv}).

\subsection{\hiConvName}
\label{sec:octaconv}

\begin{figure*}[th!]
\vspace{-2em}
\begin{center}
\resizebox{\textwidth}{!}{
    \hspace{.02\textwidth}
    \begin{subfigure}[b]{.5\textwidth}
    \centering
      \includegraphics[width=\textwidth]{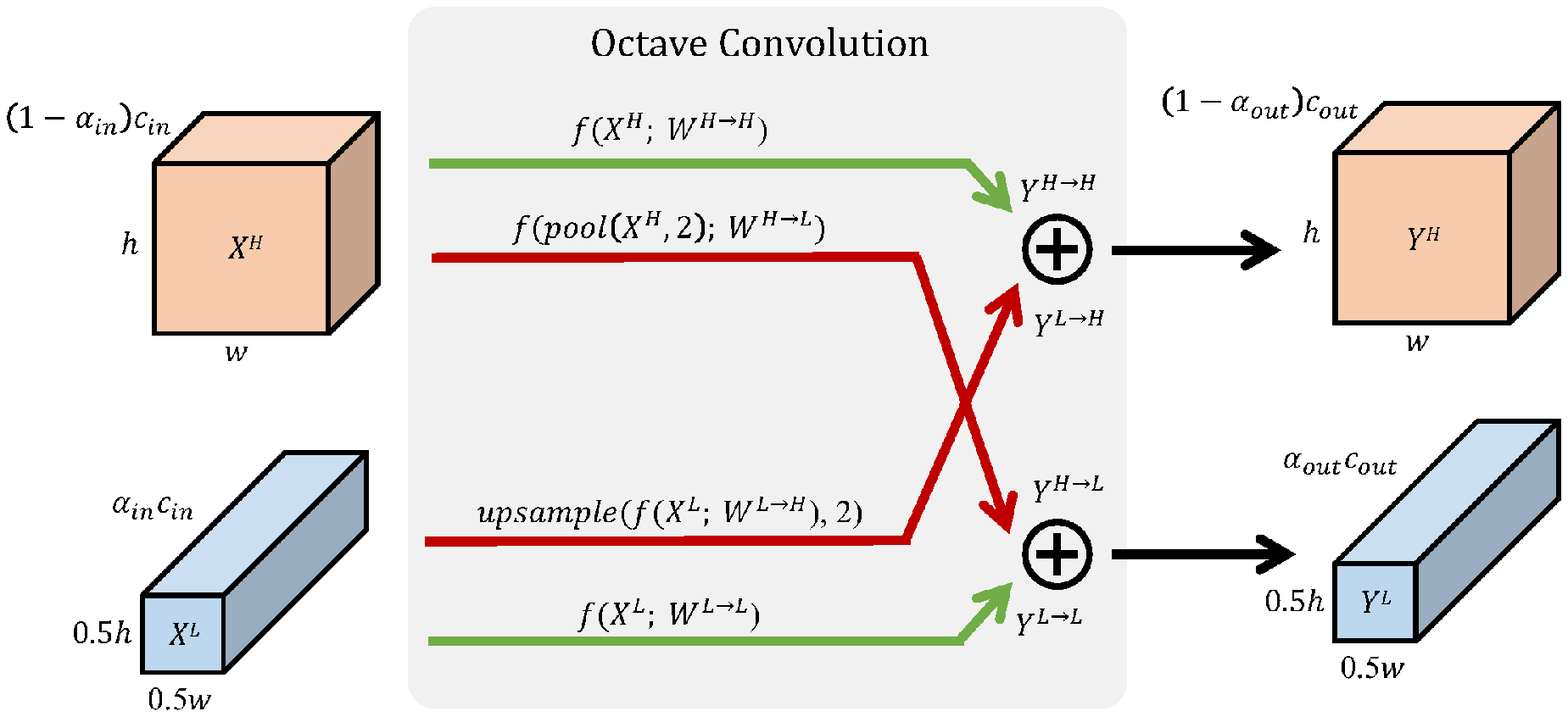}
      \caption{Detailed design of the \hiConvName. Green arrows correspond to information updates while red arrows facilitate information exchange between the two frequencies.}
    \end{subfigure}
    \hspace{.06\textwidth}
    \begin{subfigure}[b]{.4\textwidth}
    \centering
      \hspace{-18pt}
      \includegraphics[width=.5\textwidth]{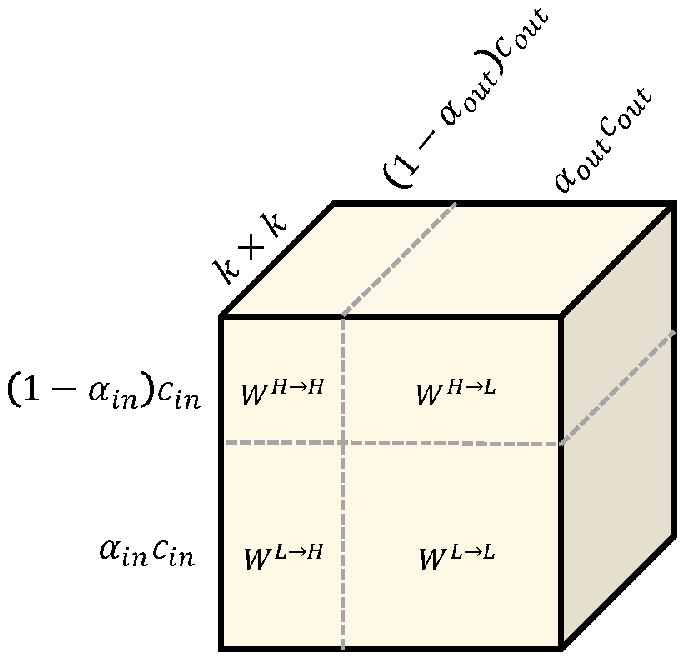}
      \vspace{-1pt}
      \caption{The \hiConvName kernel.  The $k \times k$ \hiConvName kernel $W \in \mathbb{R}^{c_{in}\times c_{out} \times k \times k}$ is equivalent to the vanilla convolution kernel in the sense that the two have the exact same number of parameters.}
    \end{subfigure}
    \hspace{.02\textwidth}
}
\end{center}
\vspace{-5mm}
   \caption{\hiConvName.  We set $\alpha_{in}=\alpha_{out}=\alpha$ throughout the network, apart from the first and last \hiConv of the network where $\alpha_{in} = 0, \alpha_{out} = \alpha$ and $\alpha_{in} = \alpha, \alpha_{out} = 0$, respectively.}
\label{fig:HiConv_details}
\end{figure*}

The octave feature representation presented in Section~\ref{sec:feature_representation} reduces the spatial redundancy and is more compact than the original representation. However, the vanilla convolution cannot directly operate on such a representation, due to differences in spatial resolution in the input features. A naive way of circumventing this is to up-sample the low-frequency part $X^L$ to the original spatial resolution, concatenate it with $X^H$ and then convolve, which would lead to extra costs in computation and memory and diminish all the savings from the compression.
In order to fully exploit our compact \hierarchicalfeaturerepresentation, we introduce  \hiConvName, which can directly operate on factorized tensors $X = \{X^H,X^L\}$ without requiring any extra computational or memory overhead.

\myparagraph{Vanilla Convolution} 
Let $W \in \mathbb{R}^{c \times \hk \times \wk}$ denote  a $\hk \times \wk$  convolution kernel and $X, Y \in \mathbb{R}^{c \times h \times w}$ denote the input and output tensors, respectively. Each feature map in $Y_{p,q} \in \mathbb{R}^{c}$
can be computed by
\begin{equation}
    Y_{p,q} = \sum_{i,j \in \mathcal{N}_k}{W_{i+\frac{\hk-1}{2},j+\frac{\wk-1}{2}}}^\top X_{p+i,q+j},
    \label{eq:conv}
\end{equation}
where $(p,q)$ denotes the location coordinate and  $\mathcal{N}_k = \{ (i,j) : i=\{-\frac{\hk-1}{2}, \ldots, \frac{\hk-1}{2} \}, j=\{-\frac{\wk-1}{2}, \ldots, \frac{\wk-1}{2} \} \}$ defines a local neighborhood. For simplicity, in all equations we omit the padding, we assume $k$ is an odd number and that the input and output data have the same dimensionality, \emph{i.e.} $c_{in}=c_{out}=c$.

\myparagraph{\hiConvName}
The goal of our design is to effectively process the low and high frequency in their corresponding frequency tensor but also enable efficient inter-frequency communication. 
Let $X,Y$ be the factorized input and output tensors. 
Then the high- and low-frequency feature maps of the output $Y=\{Y^H,Y^L\}$ will be given by $Y^H = Y^{H \rightarrow H} + Y^{L \rightarrow H}$ and $Y^L = Y^{L \rightarrow L} + Y^{H \rightarrow L}$, respectively, where $Y^{A \rightarrow B}$ denotes the convolutional update from feature map group $A$ to group $B$. Specifically, $Y^{H \rightarrow H}, Y^{L \rightarrow L}$ denote intra-frequency update, while $Y^{H \rightarrow L}, Y^{L \rightarrow H}$ denote inter-frequency communication.

To compute these terms, we split the convolutional kernel $W$ into two components $W = [W^H, W^L]$ responsible for convolving with $X^H$ and $X^L$ respectively. Each component can be further divided into intra- and inter-frequency part: $W^H=[W^{H \rightarrow H},W^{L \rightarrow H}]$ and $W^L=[W^{L \rightarrow L}, W^{H \rightarrow L}]$ with the parameter tensor shape shown in Figure~\ref{fig:HiConv_details}(b).
Specifically for high-frequency feature map, we compute it at location $(p,q)$ by using a regular convolution for the intra-frequency update, and for the inter-frequency communication we can fold the up-sampling over the feature tensor $X^L$ into the convolution, removing the need of explicitly computing and storing the up-sampled feature maps as follows:
\begin{equation}
    \begin{aligned}
    Y^H_{p,q} =& Y^{H \rightarrow H}_{p,q} + Y^{L \rightarrow H}_{p,q} \\
    =& \sum_{i,j \in \mathcal{N}_k}
    {W^{H \rightarrow H}_{i+\frac{\hk-1}{2},j+\frac{\wk-1}{2}}}^\top X^H_{p+i,q+j} \\
     & + \sum_{i,j \in \mathcal{N}_k} {W^{L \rightarrow H}_{i+\frac{\hk-1}{2},j+\frac{\wk-1}{2}}}^\top X^L_{(\lfloor \frac{p}{2} \rfloor +i),(\lfloor \frac{q}{2} \rfloor +j)},
    \end{aligned}
    \label{eq:yh}
\end{equation}
where $\lfloor \cdot \rfloor$  denotes the floor operation.
Similarly, for the low-frequency feature map, we compute the intra-frequency update using a regular convolution. Note that, as the map is in one octave lower, the convolution is also low-frequency w.r.t. the high-frequency coordinate space. For the inter-frequency communication we can again fold the down-sampling of the feature tensor $X^H$ into the convolution as follows:
\begin{equation}
    \begin{aligned}
    Y^L_{p,q} = &  Y^{L \rightarrow L}_{p,q} + Y^{H \rightarrow L}_{p,q} \\
    = & \sum_{i,j \in \mathcal{N}_k}
    {W^{L \rightarrow L}_{i+\frac{\hk-1}{2},j+\frac{\wk-1}{2}}}^\top X^L_{p+i,q+j} \\
    & + \sum_{i,j \in \mathcal{N}_k} {W^{H \rightarrow L}_{i+\frac{\hk-1}{2},j+\frac{\wk-1}{2}}}^\top X^H_{(2 * p + 0.5 +i),(2 * q + 0.5 + j)},
    \end{aligned}
    \label{eq:yl}
\end{equation}
where multiplying a factor $2$ to the locations $(p,q)$ performs down-sampling, and further shifting the location by half step is to ensure the down-sampled maps well aligned with the input. However, since the index of $X^H$ can only be an integer, we could either round the index to $(2 * p + i, 2 * q + j)$ or approximate the value at $( 2 * p + 0.5 +i,2 * q + 0.5 + j)$ by averaging all 4 adjacent locations. The first one is also known as strided convolution and the second one as average pooling. As we discuss in Section~\ref{sec:implementation} and Fig.~\ref{fig:center-shitted}, strided convolution leads to misalignment; we therefore use average pooling to approximate this value for the rest of the paper.

An interesting and useful property of the \hiConvName is the larger receptive field for the low-frequency feature maps. Convolving the low-frequency part $X^L$ with $k \times k$ convolution kernels, results in an effective enlargement of the receptive field by a factor of 2 compared to vanilla convolutions. This further helps each \hiConv layer capture more contextual information from distant locations and can potentially improve recognition performance.

%% file: tex/s4_implement.tex
\subsection{Implementation Details}
\label{sec:implementation}

As discussed in the previous subsection, the index $\{(2 * p + 0.5 +i),(2 * q + 0.5 + j)\}$ has to be an integer for Eq.~\ref{eq:yl}. Instead of rounding it to $\{(2 * p + i),(2 * q + j)\}$, \emph{i.e.} conduct convolution with stride 2 for down-sampling, we adopt average pooling to get more accurate approximation. This helps alleviate misalignments that appear when aggregating information from different scales, as shown in Appendix A. and Appendix C.. We can now rewrite the output $Y=\{Y^H,Y^L\}$ of the \hiConvName using average pooling for down-sampling as:
\begin{equation}
    \begin{aligned}
    Y^H = & f(X^H ; W^{H \rightarrow H}) + \mathtt{upsample}( f(X^L ; W^{L \rightarrow H}),2) \\
    Y^L = & f(X^L ; W^{L \rightarrow L}) + f(\mathtt{pool}(X^H, 2) ; W^{H \rightarrow L}) ) ,
    \end{aligned}
    \label{eq:hiconv_implement}
\end{equation}
where $f(X; W)$ denotes a convolution with parameters $W$, $\mathtt{pool}(X, k)$ is an average pooling operation with kernel size $k \times k$ and stride $k$. $\mathtt{upsample}(X, k)$ is an up-sampling operation by a factor of $k$ via nearest interpolation. 

The details of the \hiConv operator implementation are shown in Figure~\ref{fig:HiConv_details}. It consists of four computation paths that correspond to the four terms in Eq.~(\ref{eq:hiconv_implement}): two green paths correspond to information updating for the high- and low-frequency feature maps, and two red paths facilitate information exchange between the two octaves.

\myparagraph{Group and Depth-wise convolutions}
The \hiConvName can also be adopted to other popular variants of the vanilla convolution such as group~\cite{ResNeXt} or depth-wise~\cite{MobileNetV1} convolutions. For the group convolution case, we simply set all four convolution operations that appear inside the design of the \hiConv to group convolutions.
Similarly, for the depth-wise convolution case, the convolution operations are depth-wise and therefore the information exchange paths are eliminated, leaving only two depth-wise convolution operations. We note that both group \hiConv and depth-wise \hiConv reduce to their respective vanilla versions if we do not compress the low-frequency part.

\begin{table}[t]
\centering
\renewcommand{\arraystretch}{1.2}
\resizebox{1.0\columnwidth}{!}{
  \begin{tabular}{c|ccccccc}
  \toprule
   ratio ($\alpha$) &     .0    &    .125   &     .25   &    .50    &    .75    &    .875   &    1.0      \\
  \midrule
  \#FLOPs Cost      &  $100\%$  &   $82\%$  &   $67\%$  &   $44\%$  &   $30\%$  &   $26\%$  &   $25\%$    \\
  Memory Cost       &  $100\%$  &   $91\%$  &   $81\%$  &   $63\%$  &   $44\%$  &   $35\%$  &   $25\%$    \\
  \bottomrule
  \end{tabular}
}
\vspace{-1pt}
\caption{Relative theoretical gains for the proposed \hierarchicalfeaturerepresentation over vanilla feature maps for varying choices of the ratio $\alpha$ of channels used by the low-frequency feature. When $\alpha = 0$, no low-frequency feature is used which is the case of vanilla convolution. 
}
\label{tab:hiconv:resources-cost}
\end{table}

\myparagraph{Efficiency analysis}
Table~\ref{tab:hiconv:resources-cost} shows the theoretical computational cost and memory consumption of \hiConv over the vanilla convolution and vanilla feature map representation. More information on deriving the theoretical gains presented in Table~\ref{tab:hiconv:resources-cost} can be found in the supplementary material. We note the theoretical gains are calculated per convolutional layer. 
In Section~\ref{sec:experiments} we present the corresponding practical gains on real scenarios and show that our \hiConv implementation can sufficiently approximate the theoretical numbers.

\myparagraph{Integrating \hiConv into backbone networks}
\hiConv is backwards compatible with vanilla convolution and can be inserted to regular convolutional networks without special adjustment. To convert a vanilla feature representation to a \hierarchicalfeaturerepresentation, \ie at the first \hiConv layer, we set $\alpha_{in} = 0$ and $\alpha_{out} = \alpha$. In this case, \hiConv paths related to the low-frequency input is disabled, resulting in a simplified version which only has two paths. To convert the \hierarchicalfeaturerepresentation back to vanilla feature representation, \ie at the last \hiConv layer, we set $\alpha_{out} = 0$. In this case, \hiConv paths related to the low-frequency output is disabled, resulting in a single full resolution output.

\label{sec:imp:mg-conv}
\myparagraph{Comparison to Multi-grid Convolution~\cite{ke2017multigrid}}
The multigrid conv (MG-Conv)~\cite{ke2017multigrid} is a bi-directional and cross-scale convolution operator. Though being conceptually similar, our \hiConv is different from MG-Conv in both the core motivation and design. MG-Conv aims to exploit multi-scale information in existing CNNs, while \hiConv is focusing on reducing spatial redundancy among neighborhood pixels. In terms of design, MG-Conv adopts max-pooling for down-sampling. This requires extra memory for storing the index of the maximum value during training and further yields lower accuracy (see Appendix C.). MG-Conv also first up-samples and then convolves with the enlarged feature maps. Differently, \hiConv aims for reducing spatial redundancy and is a naive extension of convolution operator. It uses average pooling to distill low-frequency features without extra memory cost and its upsampling operation follows the convolution, and is thus more efficient than MG-Conv. 
The meticulous design of the lateral paths are essential for \hiConv to be much more memory and computationally efficient than MG-Conv and improve accuracy without increasing the network complexity. We compare \hiConv to MG-Conv in Table~\ref{tab:cls:sota:medium}.

%% file: tex/s5_exp.tex
\section{Experimental Evaluation}
\label{sec:experiments}

In this section, we validate the effectiveness and efficiency of the proposed \hiConvName for both 2D and 3D networks. We first present ablation studies for image classification on ImageNet~\cite{imagenet} and then compare it with the state-of-the-art. Then, we show the proposed \hiConv also works in 3D CNNs using Kinetics-400~\cite{kay2017kinetics, k400} and Kinetics-600~\cite{k600} datasets. The best results per category/block are highlighted in bold font throughout the paper.

\subsection{Experimental Setups}

\myparagraph{Image classification} We examine \hiConv on a set of most popular CNNs~\cite{MobileNetV1,MobileNetV2,ResNetV1,ResNetV2,densenet,ResNeXt,SENet} by replacing the regular convolutions with \hiConv (except the first convolutional layer before the max pooling). The resulting networks only have one global hyper-parameter $\alpha$, which denotes the ratio of low frequency part. We do apple-to-apple comparison and reproduce all baseline methods by ourselves under the same training/testing setting for internal ablation studies. All networks are trained with na\"ive softmax cross entropy loss except that the MobileNetV2 also adopts the label smoothing~\cite{InceptionV4}, and the best ResNet-152 adopts both label smoothing and mixup~\cite{mixup} to prevent overfitting. Same as \cite{bLNet}, all networks are trained from scratch and optimized by SGD with cosine learning rate~\cite{imnet1hour}. Standard accuracy of single centeral crop~\cite{ResNetV1,ResNetV2,ResNeXt,bLNet,ELASTIC} on validation set is reported.

\myparagraph{Video action recognition}
We use both Kinetics-400 \cite{kay2017kinetics, k400} and Kinetics-600 \cite{k600} for human action recognition. We choose standard baseline backbones from Inflated 3D ConvNet~\cite{nonlocal} and compare them with the \hiConv counterparts. We follow the setting from \cite{nonlocal-git} using frame length of 8 as standard input size,  training 300k iterations in total, and averaging the predictions over 30 crops during inference time. To make fair comparison, we report the performance of the baseline and \hiConv under precisely the same settings.

\input{tex/s5_exp_image_cls.tex}

\input{tex/s5_exp_video_cls.tex}

%% file: tex/s5_exp_image_cls.tex
\subsection{Ablation Study on ImageNet}
\label{sec:exp:imgcls}
\label{sec:exp:imgcls:ablation}

We conduct a series of ablation studies aiming to answer the following questions: 
1) \textit{Does \hiConv have better FLOPs-Accuracy trade-off than vanilla convolution? }
2) \textit{In which situation does the \hiConv work the best?}

\begin{figure}[t]
\vspace{-1.5em}
\centering
\hspace{-14pt}
    \resizebox{1.05\columnwidth}{!}{
        \input{tex/s5_exp_image_ablation_figure}
    }\vspace{-8pt}
    \caption{
    Ablation study results on ImageNet. \hiConv-equipped models are more efficient and accurate than baseline models. Markers in black in each line denote the corresponding baseline models without \hiConv. The colored numbers are the ratio $\alpha$. Numbers in X axis denote FLOPs in logarithmic scale.}
\label{fig:imnet:ablation}
\end{figure}
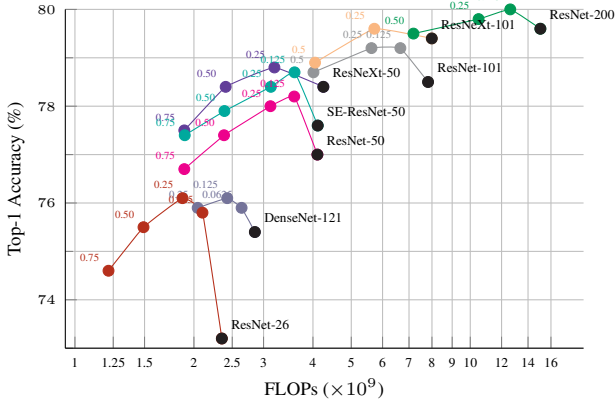

\myparagraph{Results on ResNet-50}
We begin with using the popular ResNet-50~\cite{ResNetV2} as the baseline CNN and replacing the regular convolution with our proposed \hiConv to examine the flops-accuracy trade-off. In particular, we vary the global ratio $\alpha \in \{0.125, 0.25, 0.5, 0.75\} $ to compare the image classification accuracy versus computational cost (\empty{i.e.} FLOPs)~\cite{ResNetV1,ResNetV2,ResNeXt,DPN} with the baseline. The results are shown in Figure~\ref{fig:imnet:ablation} in pink.

We make following observations. 1) The flops-accuracy trade-off curve is a concave curve, where the accuracy first rises up and then slowly goes down. 2) We can see two sweet spots:  The first at $\alpha=0.5$, where the network gets similar or better results even when the FLOPs are reduced by about half;
the second at $\alpha=0.125$, where the network reaches its best accuracy, 1.2\% higher than baseline (black circle). We attribute the increase in accuracy to \hiConv's effective design of multi-frequency processing and the corresponding enlarged receptive field which provides more contextual information to the network. While reaching the accuracy peak at $0.125$, the accuracy does not suddenly drop but decreases slowly for higher ratios $\alpha$, indicating reducing the resolution of the low frequency part does not lead to significant information loss. Interestingly, $75\%$ of the feature maps can be compressed to half the resolution with only $0.3\%$ accuracy drop, which demonstrates effectiveness of grouping and compressing the smoothly changed feature maps for reducing the spatial redundancy in CNNs. In Table~\ref{tab:hiconv:res50-cpu} we demonstrate the theoretical FLOPs saving of \hiConv is also reflected in the actual CPU inference time in practice. For ResNet-50, we are close to obtaining theoretical FLOPs speed up. These results indicate \hiConv is able to deliver important practical benefits, rather than only saving FLOPs in theory.

\begin{table}[t]
\vspace{-1.5em}
\centering
\renewcommand{\arraystretch}{1.2}
\resizebox{1.0\columnwidth}{!}{
\begin{tabular}{cccccc}
   \toprule
    ratio ($\alpha$) & Top-1 (\%) & \#FLOPs (G) &      Inference Time (ms)          & Backend      \\ 
   \midrule
      N/A   &    77.0    &  4.1 &    119         &  MKLDNN       \\ 
      N/A   &    77.0    &  4.1 &    115    &  TVM          \\
     .125 &    78.2    &  3.6 &    116    &  TVM          \\ 
     .25  &    78.0    &  3.1 &   ~~99     &  TVM          \\ 
     .5   &    77.4    &  2.4 &   ~~74    &  TVM          \\ 
     .75  &    76.7    &  1.9 &   ~~61    &  TVM          \\ 
  \bottomrule
\end{tabular}
}
\vspace{-8pt}
\caption{Results of ResNet-50. Inference time is measured on Intel Skylake CPU at 2.0 GHz (single thread). We report Intel(R) Math Kernel Library for Deep Neural Networks v0.18.1 (MKLDNN)~\cite{mkldnn} inference time for vanila ResNet-50. Because vanilla ResNet-50 is well optimized by Intel, we also show MKLDNN results as additional performance baseline. \hiConv networks are compiled by TVM~\cite{tvm} v0.5.}
\label{tab:hiconv:res50-cpu}
\end{table}

\myparagraph{Results on more CNNs}
To further examine if the proposed \hiConv works for other networks with different depth/wide/topology, we select the currently most popular networks as baselines and repeat the same ablation study. These networks are ResNet-(26;50;101;200)~\cite{ResNetV2}, ResNeXt-(50,32$\times$4d;101,32$\times$4d)~\cite{ResNeXt}, DenseNet-121~\cite{densenet} and SE-ResNet-50~\cite{SENet}. The ResNeXt is chosen for assessing the \hiConv on group convolution, while the SE-Net~\cite{SENet} is used to check if the gain of SE block found on vanilla convolution based networks can also be seen on \hiConv. As shown in Figure~\ref{fig:imnet:ablation}, \hiConv equipped networks for different architecture behave similarly to the \hiConvPrefix-ResNet-50, where the FLOPs-Accuracy trade-off is in a concave curve and the performance peak also appears at ratio $\alpha=0.125$ or $\alpha=0.25$. The consistent performance gain on a variety of backbone CNNs confirms that \hiConv is a good replacement of vanilla convolution.

\begin{figure}[t]
\centering
\vspace{-3mm}
\includegraphics[height=1.4cm,width=7.5cm]{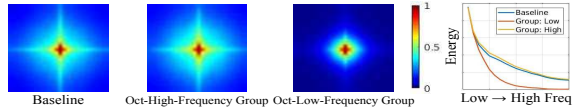}
\vspace{-1mm}
\caption{Frequency analysis for activation maps in different groups. `Baseline` refers to vanilla ResNet. 10k activation maps are sampled from ResNet-101(Res3).}
\label{fig:freq}
\end{figure}

\myparagraph{Frequency Analysis}
Figure~\ref{fig:freq} shows the frequency analysis results. We conducted the Fourier transform for each group of feature maps and visualized the averaged results. From the energy map, the low frequency group does not contain high frequency signal, while the high frequency group contains both low and high frequency signals. This confirms that low-frequency group indeed captures low-frequency information as expected. Note that OctConv gives the high frequency group the flexibly to store both low and high frequency signals for better learning capacity.

\myparagraph{Summary}
1) \hiConv can help CNNs improve the accuracy while decreasing the FLOPs, deviating from other methods that reduce the FLOPs with a cost of lower accuracy. 2) At test time, the gain of \hiConv over baseline models increases as the test image resolution grows because \hiConv can detect large objects better due to its larger receptive field, see Appendix C. 3) Both the information exchanging paths are important, since removing any of them can lead to accuracy drop, see Appendix C. 4) Shallow networks, \emph{e.g.} ResNet-26, have a rather limited receptive field, and can especially benefit from \hiConv, which greatly enlarges their receptive field.

\subsection{Comparing with SOTAs on ImageNet}
\label{sec:exp:imgcls:sota}

\begin{table}[t]
\centering
\renewcommand{\arraystretch}{1.2}
\resizebox{\columnwidth}{!}{
  \begin{tabular}{lcccccc}
   \toprule
   Method          & ratio ($\alpha$) &  \#Params (M) &       \#FLOPs (M)  &   CPU (ms)   &    Top-1 (\%)  \\
   \midrule
   CondenseNet ($G = C = 8$)~\cite{CondenseNet}
                   &   -   &     2.9  &         274  &    -    &    71.0        \\
   1.5 ShuffleNet (v1)~\cite{ShuffleNetV1}
                   &   -   &     3.4  &         292   &    -    &    71.5        \\
   1.5 ShuffleNet (v2)~\cite{ShuffleNetV2} 
                   &   -   &     3.5  &         299   &    -    & \textbf{72.6}  \\
   \midrule
   0.75 MobileNet (v1)~\cite{MobileNetV1}
                   &  -    &     2.6  &         325   &  13.4 & ~~70.3$^*$     \\ 
   0.75 \hiConvPrefix-MobileNet (v1) (ours)
                   & .375  &     2.6  & \textbf{213} &  \textbf{11.9} &   70.5        \\
   1.0 \hiConvPrefix-MobileNet (v1) (ours)
                   & .5    &     4.2  &         321   & 18.4  & \textbf{72.5}  \\
   \midrule
   1.0 MobileNet (v2)~\cite{MobileNetV2}
                   &  -    &     3.5  &         300  & 24.5 &    72.0        \\
   1.0 \hiConvPrefix-MobileNet (v2) (ours)
                   & .375  &     3.5  & \textbf{256} & \textbf{17.1} &    72.0        \\
   1.125 \hiConvPrefix-MobileNet (v2) (ours)
                   & .5    &     4.2  &         295  & 26.3 & \textbf{73.0}  \\
  \bottomrule
  \end{tabular}
}
\vspace{-8pt}
\caption{ImageNet classification results for \textit{Small} models. $^*$ indicates it is better than original reproduced by MXNet GluonCV v0.4~\cite{gluoncvnlp2019}. The inference speed is tested using TVM on Intel Skylake processor (2.0GHz, single thread)\protect\footnotemark.}
\label{tab:cls:sota:small}
\end{table}

\input{tex/s5_exp_image_cls_medium_table.tex}

\input{tex/s5_exp_image_cls_large_table.tex}


\myparagraph{Small models}
We adopt the most popular light weight networks as baselines and examine if \hiConv works well on these compact networks with depth-wise convolution. In particular, we use the ``0.75 MobileNet (v1)''~\cite{MobileNetV1} and ``1.0 MobileNet (v2)''~\cite{MobileNetV2} as baseline and replace the regular convolution with our proposed \hiConv. The results are shown in Table~\ref{tab:cls:sota:small}. We find that \hiConv can reduce the FLOPs of MobileNetV1 by $34\%$, and provide better accuracy and faster speed in practice; it is able to reduce the FLOPs of MobileNetV2 by 15\%, achieving the same accuracy with faster speed. When the computation budget is fixed, one can adopt wider models to increase the learning capacity because \hiConv can compensate the extra computation cost. In particular, our \hiConv equipped networks achieve $2\%$ improvement on MobileNetV1 under the same FLOPs and 1\% improvement on MobileNetV2. \footnotetext[2]{For small models, we should notice according to arithmetic intensity \cite{intensity}, real execution time is not only bounded by FLOPS. }

\myparagraph{Medium models}
In the above experiment, we have compared and shown that \hiConv is complementary with a set of state-of-the-art CNNs~\cite{ResNetV1,ResNetV2,ResNeXt,densenet,MobileNetV1,MobileNetV2,SENet}. In this part, we compare \hiConv with MG-Conv~\cite{ke2017multigrid}, GloRe~\cite{glorea}, Elastic~\cite{ELASTIC} and bL-Net~\cite{bLNet} which share a similar idea as our method. Seven groups of results are shown in Table~\ref{tab:cls:sota:medium}. In group 1, our \hiConvPrefix-ResNet-26 shows $0.6\%$ better accuracy than R-MG-34 while costing only one third of FLOPs and half of \#Params. Also, our \hiConvPrefix-ResNet-50, which costs less than half of FLOPS, achieves $1.9\%$ higher accuracy than R-MG-34. In group 2, adding our \hiConv to GloRe network reduces the FLOPs with better accuracy. In group 3, our \hiConvPrefix-ResNeXt-50 achieves better accuracy than the Elastic~\cite{ELASTIC} based method (78.8\% v.s. 78.4\%) while reducing the computational cost by 31\%. In group 4, the \hiConvPrefix-ResNeXt-101 also achieves higher accuracy than the Elastic based method (79.6\% v.s. 79.2\%) while costing 38\% less computation. When compared to the bL-Net~\cite{bLNet}, \hiConv equipped methods achieve better FLOPs-Accuracy trade-off without bells and tricks. When adopting the tricks used in the baseline bL-Net~\cite{bLNet}, our \hiConvPrefix-ResNet-50 achieves 0.9\% higher accuracy than bL-ResNet-50 under the same computational budget (group 5), and \hiConvPrefix-ResNeXt-50 (group 6) and \hiConvPrefix-ResNeXt-101 (group 7) get better accuracy under comparable or even lower computational budget. This is because MG-Conv~\cite{ke2017multigrid}, Elastic-Net~\cite{ELASTIC} and bL-Net~\cite{bLNet} are designed following the principle of introducing multi-scale features without considering reducing the spatial redundancy. In contrast, \hiConv is born for solving the high spatial redundancy problem in CNNs, uses more efficient strategies to store and process the information throughout the network, and can thus achieve better efficiency and performance.

\myparagraph{Large models}
Table~\ref{tab:cls:sota:large} shows the results of \hiConv in large models. Here, we choose the ResNet-152 as the backbone CNN, replacing the first $7\times7$ convolution by three $3\times3$ convolution layers and removing the max pooling by a lightweight residual block~\cite{bLNet}. We report results for \hiConvPrefix-ResNet-152 with and without the SE-block~\cite{SENet}. As can be seen, our \hiConvPrefix-ResNet-152 achieves accuracy comparable to the best manually designed networks with less FLOPs (10.9G v.s.~12.7G). Since our model does not use group or depth-wise convolutions, it also requires significantly less GPU memory, and runs faster in practice compared to the SE-ShuffleNet v2-164 and AmoebaNet-A (N=6, F=190) which have low FLOPs in theory but run slow in practice due to the use of group and depth-wise convolutions. Our proposed method is also complementary to Squeeze-and-excitation~\cite{SENet}, where the accuracy can be further boosted when the SE-Block is added (last row). 

%% file: tex/s5_exp_image_ablation_figure.tex
\tikzset{every mark/.append style={solid}}
\pgfplotsset{ grid=both, width=\columnwidth, try min ticks=5,
	 every axis x label/.style={at={(ticklabel cs:0.5)},anchor=north},
	 every axis y label/.style={at={(ticklabel cs:0.5)},rotate=90,anchor=south},
	 every axis/.append style={font=\scriptsize,thick,mark=*,smooth,tension=0.18},
	 legend cell align=left, legend style={fill opacity=0.8},
	 set layers,
     mark layer=axis foreground,
     mmark/.style={mark=*,solid},
     dash/.style={mark=o,dashed,opacity=0.6},
}

\begin{tikzpicture}
\begin{semilogxaxis}[%
    axis x line*=bottom,
    axis y line*=left,
	height=6cm,
	xlabel={FLOPs ($ \times 10^9$)},
	ylabel={Top-1 Accuracy (\%)},
	ymin=73,
	ymax=80.0,
	minor y tick num=1,
	nodes near coords,
    every node near coord/.style={anchor=south east, font=\fontsize{4}{5}\selectfont},
    point meta=explicit symbolic,
    nodes near coords,
    mark size=2pt, 
    xtick={1,1.25,1.5,2,2.5,3,4,5,6,7,8,9,10,12,14,16},
    xticklabels={1,1.25,1.5,2,2.5,3,4,5,6,7,8,9,10,12,14,16},
    ylabel style={at={(-0.06,0.5)}}, 
    xlabel style={at={(0.5,-0.06)}},  
    every tick label/.append style={font=\tiny},
]

 	\pgfplotstableread{ 
 	 	marker  flop  acc   myanchor
        \empty{} 2.852 75.4 north
       0.0625  2.640 75.9  north
       0.125  2.428 76.1   south
       0.25   2.044 75.9   south
    }{\data} 
 	\addplot+[CadetBlue, mmark,mark options={CadetBlue}
 	    ]
 	 	table[x=flop, y=acc, meta=marker] \data; 
 	
 	\pgfplotstableread{
 	    marker         flop         acc
 	    \empty{}       2.353        73.2
        0.125		   2.102	    75.8 		
        0.25		   1.871	    76.1 		
        0.50		   1.491	    75.5 		
        0.75		   1.216	    74.6 		
    }{\data}
 	\addplot+[BrickRed, mmark,mark options={BrickRed}] 
 	    table[x=flop, y=acc, meta=marker] \data;
    
 	\pgfplotstableread{ 
 	 	marker  flop  acc
 	 	\empty{} 4.250  78.4
 	 	0.25  3.196 78.8
        0.50  2.406 78.4
        0.75  1.891 77.5
    }{\data}
 	\addplot+[RoyalPurple, mmark,mark options={RoyalPurple}] 
 	    table[x=flop, y=acc, meta=marker] \data; 

 	\pgfplotstableread{ 
 	 	marker         flop         acc
        \empty{}         4.113			77.6 
        0.125			 3.594			78.7 
        0.25			 3.130			78.4 
        0.50			 2.389			77.9 
        0.75			 1.896			77.4 
    }{\data} 	
 	\addplot+[Emerald, mmark,mark options={Emerald}]
 	 	table[x=flop, y=acc, meta=marker] \data; 

 	\pgfplotstableread{ 
 	 	marker  flop  acc   
        \empty{} 7.822  78.5 
        0.125  6.656 79.2
        0.25   5.625 79.2  
        0.5    4.012 78.7   
    }{\data}  	
 	\addplot+[Gray, mmark,mark options={Gray}] 
 	    table[x=flop, y=acc, meta=marker] \data; 
  	
 	\pgfplotstableread{ 
 	 	marker  flop  acc   
        \empty{} 7.993  79.4 
        0.25   5.719 79.6 
        0.5    4.050 78.9  
    }{\data}  	
 	\addplot+[Apricot, mmark,mark options={Apricot}] 
 	    table[x=flop, y=acc, meta=marker] \data; 
 	
 	\pgfplotstableread{ 
 	 	marker  flop  acc   
        \empty{} 15.044 79.6
        0.125 12.623 80.0
        0.25  10.497 79.8
        0.50  7.183  79.5	        
    }{\data}  	
 	\addplot+[ForestGreen, mmark,mark options={ForestGreen}] 
 	    table[x=flop, y=acc, meta=marker] \data; 
    
 	\pgfplotstableread{ 	
     	marker         flop         acc
     	\empty{}	   4.105        77.0	
        0.125		   3.587        78.2
        0.25		   3.123        78.0
        0.50		   2.383        77.4		
        0.75		   1.891        76.7		
 	}{\data}
 	\addplot+[Magenta, mmark,mark options={Magenta}] 
    	table[x=flop, y=acc, meta=marker] \data;

    \pgfplotstableread{
        baseflop     baseacc    basemarker
        2.353        73.2		ResNet-26
        4.105        77.0		ResNet-50
        4.113	     77.6		SE-ResNet-50
        4.250        78.4		ResNeXt-50
        7.822        78.5 		ResNet-101
        7.993        79.4 		ResNeXt-101
        2.852        75.4 		DenseNet-121 
        15.044       79.6		ResNet-200
    } {\basedata}
    \addplot[Black, mmark, only marks, mark options={Black},
    node near coord style={text=black, anchor=south west, font=\tiny}] table [x=baseflop, y=baseacc, meta=basemarker] \basedata;
    
\end{semilogxaxis}
\end{tikzpicture}

%% file: tex/s5_exp_image_cls_medium_table.tex
\begin{table}[t]
\centering
\renewcommand{\arraystretch}{1.2}
\resizebox{\columnwidth}{!}{
  \begin{tabular}{lcccccc}
   \toprule
   Method           & ratio ($\alpha$) &   Depth  &      \#Params (M)  &       \#FLOPs (G)  &      Top-1 (\%)   \\
   \midrule
   R-MG-34~\cite{ke2017multigrid}  & - & 34 &       32.9        & 5.8  &  75.5  \\
   \hiConvPrefix-ResNet-26 (ours)
                    &  .25  &    26    &\textbf{16.0} & \textbf{1.9} & 76.1     \\
   \hiConvPrefix-ResNet-50 (ours)
                    &  .5   &    50  &  25.6 & 2.4 &      \textbf{77.4 }        \\   
   \midrule
   \midrule
   ResNet-50 + GloRe~\cite{glorea} (+3 blocks \@ Res4)
                    &   -   &    50  &  30.5 & 5.2 &      78.4        \\   
   \hiConvPrefix-ResNet-50 (ours) + GloRe~\cite{glorea} (+3 blocks \@ Res4)
                    &  .5   &    50  &  30.5 & \textbf{3.1} &  \textbf{78.8 }   \\                     
   \midrule
   \midrule
   ResNeXt-50 + Elastic~\cite{ELASTIC}
                    &   -   &    50    &        25.2   &         4.2  &      78.4         \\
   \hiConvPrefix-ResNeXt-50 (32$\times$4d) (ours)
                    &  .25  &    50    &\textbf{25.0 } & \textbf{3.2 } & \textbf{78.8}     \\
   \midrule
   ResNeXt-101 + Elastic~\cite{ELASTIC}
                                &   -   &   101    &        44.3   &         7.9   &      79.2         \\
   \hiConvPrefix-ResNeXt-101 (32$\times$4d) (ours)
                    &  .25  &   101    &\textbf{44.2 } & \textbf{5.7} & \textbf{79.6}    \\
   \midrule
   \midrule
   bL-ResNet-50$^\ddag$ ($\alpha=4, \beta=4$) ~\cite{bLNet}
                                 &   -   &  50 (+3) &        26.2   &         2.5   &      76.9         \\
   \hiConvPrefix-ResNet-50$^\ddag$ (ours)
                    &  .5   &  50 (+3) &        25.6   &         2.5   & \textbf{77.8}   \\
   \hiConvPrefix-ResNet-50 (ours)
                    &  .5   &    50    &\textbf{25.6} & \textbf{2.4} &      77.4         \\ 
   \midrule
   bL-ResNeXt-50$^\ddag$ (32$\times$4d)~\cite{bLNet}
                                 &   -   &  50 (+3) &        26.2   &         3.0   &      78.4         \\
   \hiConvPrefix-ResNeXt-50$^\ddag$ (32$\times$4d) (ours)
                    &  .5   &  50 (+3) &        25.1   &         2.7   & \textbf{78.6}     \\
   \hiConvPrefix-ResNeXt-50 (32$\times$4d) (ours)
                    &  .5   &    50    &\textbf{25.0 } & \textbf{2.4 } &      78.4         \\
   \midrule
   bL-ResNeXt-101$^\ddag$ $^\S$ (32$\times$4d)~\cite{bLNet} 
                                 &   -   & 101 (+1) &43.4  &         4.1   &      78.9         \\
   \hiConvPrefix-ResNeXt-101$^\ddag$ $^\S$ (32$\times$4d) (ours)
                    &  .5   & 101 (+1) &        \textbf{40.1}   &         4.2  & \textbf{79.4}     \\
   \hiConvPrefix-ResNeXt-101$^\ddag$ (32$\times$4d) (ours)
                    &  .5   & 101 (+1) &        44.2   &         4.2   &      79.1         \\
   \hiConvPrefix-ResNeXt-101 (32$\times$4d) (ours)
                    &  .5   &   101    &        44.2   & \textbf{4.0} &      78.9         \\
  \bottomrule
  \end{tabular}
}
\vspace{-8pt}
\caption{ImageNet Classification results for \textit{Middle} sized models. 
$^\ddag$ refers to method that replaces ``Max Pooling'' by extra convolution layer(s)~\cite{bLNet}. $^\S$ refers to method that uses balanced residual block distribution~\cite{bLNet}.
}
\label{tab:cls:sota:medium}
\end{table}

%% file: tex/s5_exp_image_cls_large_table.tex
\begin{table*}[t]
\vspace{-1em}
\centering
\renewcommand{\arraystretch}{1.3}
\resizebox{\textwidth}{!}{
  \begin{tabular}{>{\raggedright}p{6.0cm}|>{\centering}p{1.5cm}|
  >{\centering}p{2cm}>{\centering}p{2.5cm}>{\centering}p{1.5cm}|
  >{\centering}p{1.5cm}>{\centering}p{1.5cm}>{\centering}p{1.5cm}|
  >{\centering}p{1.5cm}>{\centering}p{1.5cm}c}
  \toprule
  \multicolumn{1}{l|}{\multirow{2}{*}{Method}}
  & \multicolumn{1}{c|}{\multirow{2}{*}{\#Params (M)}}
  & \multicolumn{3}{c|}{\multirow{1}{*}{Training}} 
  & \multicolumn{3}{c|}{\multirow{1}{*}{Testing ($224 \times 224$)}} 
  & \multicolumn{3}{c}{\multirow{1}{*}{Testing ($320 \times 320$~~/~\@$331 \times 331$)}} \\
  \multicolumn{1}{l|}{}
  & \multicolumn{1}{c|}{}
  & \multicolumn{1}{c}{Input Size}
  & \multicolumn{1}{c}{Memory Cost (MB)}
  & \multicolumn{1}{c|}{Speed (im/s)}
  & \multicolumn{1}{c}{\#FLOPs (G)}& \multicolumn{1}{c}{Top-1 (\%)} & \multicolumn{1}{c|}{Top-5 (\%)}
  & \multicolumn{1}{c}{\#FLOPs (G)}& \multicolumn{1}{c}{Top-1 (\%)} & \multicolumn{1}{c}{~~~Top-5 (\%)~~} \\
   \midrule
   NASNet-A (N=6, F=168)~\cite{NASNet} 	$^\Diamond$
                   &  88.9  & \multirow{5}{*}{\makecell{$331 \times 331$ \\ / $320 \times 320$}}
                                     &$>32,480$ & 43 $^\ddagger$&   -    &  -   &  -   & 23.8  & 82.7 & 96.2 \\
   AmoebaNet-A (N=6, F=190)~\cite{AmoebaNet} 	$^\Diamond$
                   &  86.7  &       &$>32,480$ & 47 $^\ddagger$&   -    &  -   &  -   & 23.1  & 82.8 & 96.1 \\
   PNASNet-5 (N=4, F=216)~\cite{PNASNet} 	$^\Diamond$
                   &  86.1  &       &$>32,480$ & 38 $^\ddagger$&   -    &  -   &  -   & 25.0  & 82.9 & 96.2 \\
   Squeeze-Excite-Net~\cite{SENet}
                   & 115.1  &       &$>32,480$ & 43 $^\dagger$&   -    &  -   &  -   & 42.3  & 83.1 & 96.4 \\
   AmoebaNet-A (N=6, F=448)~\cite{AmoebaNet} 	$^\Diamond$
                   &   469  &       &$>32,480$ &      15 $^\S$&   -    &  -   &  -   &  104  & 83.9 & 96.6 \\           
   \midrule
   Dual-Path-Net-131~\cite{DPN}
                   &  79.5  & \multirow{5}{*}{$224 \times 224$}
                                     & 31,844  &         83~~~~  & 16.0  & 80.1 & 94.9 & 32.0  & 81.5 & 95.8 \\
   SE-ShuffleNet v2-164~\cite{ShuffleNetV2}
                   &  69.9  &       &$>32,480$ & 70 $^\dagger$& 12.7  & 81.4 &  -   &    -   &  -   &  -   \\
   Squeeze-Excite-Net~\cite{SENet}
                   & 115.1  &       & 28,696 &       78~~~  &   21  & 81.3 & 95.5 & 42.3  & 82.7 & 96.2 \\
    \textbf{\hiConvPrefix-ResNet-152}, $\alpha = 0.125$ (ours)
                   &\tbf{60.2 }&       &\tbf{15,566 }&\tbf{162 }~~~&\tbf{10.9 } & 81.4 & 95.4 &\tbf{22.2 }& 82.3 & 96.0 \\
   \textbf{\hiConvPrefix-ResNet-152 + SE}\protect{\footnotemark}, $\alpha = 0.125$ (ours) 
                   &  66.8  &       & 21,885  &     95~~~ & \textbf{10.9 } &\tbf{81.6}&\tbf{95.7}& \textbf{22.2 } &\tbf{82.9}&\tbf{96.3} \\      
  \bottomrule
  \end{tabular}
}
\vspace{-4pt}
\caption{ImageNet Classification results for \textit{Large} models. The names of \hiConv-equiped models are in bold font and performance numbers for related works are copied from the corresponding papers. Networks are evaluated using CuDNN v10.0\protect\footnotemark in flop16 on a \textit{single} Nvidia Titan V100 (32GB) for their training memory cost and speed. Works that employ neural architecture search are denoted by ($^\Diamond$). We set batch size to 128 in most cases, but had to adjust it to 64 (noted by $^\dagger$), 32 (noted by $^\ddagger$) or 8 (noted by $^\S$) for networks that are too large to fit into GPU memory.}
\label{tab:cls:sota:large}
\end{table*}

%% file: tex/s5_exp_video_cls.tex
\subsection{Experiments of Video Recognition on Kinetics}
\label{sec:exp:video}

In this subsection, we evaluate the effectiveness of \hiConv for action recognition in videos and demonstrate that our spatial \hiConv is sufficiently generic to be integrated into 3D convolution to decrease \#FLOPs and increase accuracy at the same time.
As shown in Table~\ref{tab:video:kinetics600}, \hiConv consistently decreases FLOPs and meanwhile improves the accuracy when added to C2D and I3D \cite{nonlocal,nonlocal-git}, and is also complementary to the Non-local~\cite{nonlocal}.
This is observed for models pre-trained on ImageNet \cite{imagenet} as well as models trained from scratch on Kinetics. The higher accuracy, lower FLOPs and the ability of being complimentary to existing metods, \emph{e.g.} Non-local method, confirm the effectiveness of the proposed \hiConv method. Performance further increases when combining \hiConv with the SlowFast Networks~\cite{SlowFast}. Specifically, we apply \hiConv on the spatial dimensions and SlowFast on the temporal dimension. 

\begin{table}[t]
\centering
\renewcommand{\arraystretch}{1.3}
  \resizebox{1.0\columnwidth}{!}{
   \begin{tabular}{lccc@{}l@{}}
   
   \toprule
  \multicolumn{1}{l}{Method}   & \multicolumn{1}{c}{ImageNet Pretrain}    & \multicolumn{1}{c}{\#FLOPs (G)}  & Top-1 (\%) \\
  \midrule
  \multicolumn{5}{c}{(a) Kinetics-400 \cite{k400}}\\
  \midrule
        I3D
            &           &         28.1         & 72.6   \\
        \hiConvPrefix-I3D, $\alpha$=0.1,  (ours)
            &         &           25.6       
            &   \textbf{73.6} & \textbf{(+1.0)} \\
        \hiConvPrefix-I3D, $\alpha$=0.2,  (ours)
            &       &           22.1        
            &   73.1 & (+0.5) \\
        \hiConvPrefix-I3D, $\alpha$=0.5,  (ours)
            &       &           \textbf{15.3}   
            &   72.1 & (-0.5) \\
    \midrule
         C2D
            &     \checked      &         19.3           & 71.9  \\
        \hiConvPrefix-C2D, $\alpha$=0.1,  (ours)
            &      \checked     &         \textbf{17.4} 
            & \textbf{73.8} &\textbf{(+1.9)} \\
    \midrule  
        I3D 
            &        \checked     &          28.1           & 73.3  \\
        \hiConvPrefix-I3D, $\alpha$=0.1,  (ours)
            &        \checked     &          \textbf{25.6}         & \textbf{74.6}& \textbf{(+1.3)}  \\
    \midrule
        I3D + Non-local
            &     \checked     &          33.3            & 74.7  \\
        \hiConvPrefix-I3D  + Non-local, $\alpha$=0.1,  (ours)
            &        \checked     &           \textbf{28.9}
            & \textbf{75.7} & \textbf{(+1.0)}  \\
    \midrule
        SlowFast-R50~\cite{SlowFast}
            &              &         27.6 \protect{\footnotemark}          &  75.6 \\
        \hiConvPrefix-SlowFast-R50, $\alpha$=0.1,  (ours)
            &              &         \textbf{24.5}          &  \textbf{76.2} & \textbf{(+0.6)} \\
        \hiConvPrefix-SlowFast-R50, $\alpha$=0.2,  (ours)
            &              &         22.9          &  75.8 & (+0.2)\\
   
    \midrule
    \multicolumn{5}{c}{(b) Kinetics-600 \cite{k600}}\\
    \midrule
        I3D 
                &      \checked     &        28.1           & 74.3  \\
        \hiConvPrefix -I3D, $\alpha$=0.1,  (ours)
                &      \checked     &        \textbf{25.6}          &  \textbf{76.0} & \textbf{(+1.7)}  \\
   \bottomrule
   \end{tabular}
  }
\caption{Action Recognition in videos, ablation study, all models with ResNet50 \cite{ResNetV1}.}
\label{tab:video:kinetics600}
\vspace{-6pt}
\end{table}

\footnotetext[3]{The auto-tune is set to \textit{off} when evaluating the memory cost for more accurate result, and is set to \textit{on} when measuring speed for fastest speed.}
\footnotetext[4]{An extra BatchNorm is added at the beginning of each residual function, otherwise the gradient will easily diverged due to the newly added SE module. This costs more memory and slows down the speed but can provide higher accuracy.}
\footnotetext[5]{Note that \cite{SlowFast} reports 36.1 GFLOPs at a spatial size of $256^2$, while we report (training) GFLOPs at $224^2$ for all methods.}

%% file: tex/s7_conclusion.tex
\section{Conclusion}
In this work, we address the problem of reducing spatial redundancy that widely exists in vanilla CNN models, and propose a novel \hiConvName operation to store and process low- and high-frequency features separately to improve the model efficiency. 
\hiConvName is sufficiently generic to replace the regular convolution operation in-place, and can be used in most 2D and 3D CNNs without model architecture adjustment. Beyond saving a substantial amount of computation and memory, \hiConvName can also improve the recognition performance by effective communication between the low- and high-frequency and by enlarging the receptive field size which contributes to capturing more global information. Our extensive experiments on image classification and video action recognition confirm the superiority of our method for striking a much better trade-off between  recognition performance and model efficiency, not only in FLOPs, but also in practice. 

\myparagraph{Acknowledgement}
We would like to thank Min Lin and Xin Zhao for helpful discussions of the code development.

%% file: supp/content.tex
\input{supp/t0_misalignment.tex}

\input{supp/t1_flops-mem-cost.tex}

\input{supp/t2_imnet_results.tex}

%% file: supp/t0_misalignment.tex
\section*{Appendix A. The Misalignment Problem}

\begin{figure}[h]
\centering
\resizebox{\columnwidth}{!}{
	\includegraphics[]{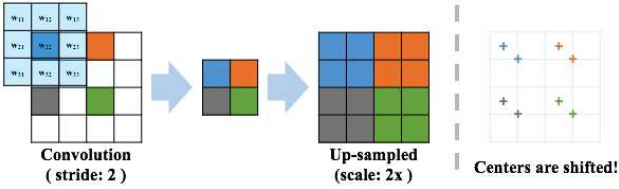}
}
\caption{Strided convolution may cause misaligned feature maps after up-sampling.}
\label{fig:center-shitted}
\end{figure}

As shown in Figure~\ref{fig:center-shitted}, up-sampling after the strided convolution with odd convolutional filter, \emph{e.g.} $3\times3$, will cause the entire feature map to move to the lower right, which is problematic when we add the up-sampled shifted map with the unshifted map.

%% file: supp/t1_flops-mem-cost.tex
\section*{Appendix B. Relative Theoretical Gains of \hiConv}
In Table 1 of the main paper, we reported the relative theoretical gains of the proposed multi-frequency feature representation over regular feature representation with respect to memory footprint and computational cost, as measured in FLOPS (\ie multiplications and additions). In this section, we show how the gains are estimated in theory.

\begin{figure}[t!]
    \centering
    \begin{subfigure}[b]{0.4\columnwidth}
      \includegraphics[width=\textwidth]{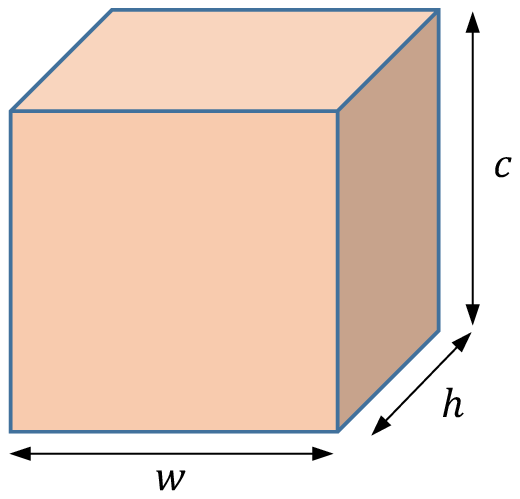}
      \caption{}
      \label{sup:fig:teaser_a}
    \end{subfigure}
    \hspace{18pt}
    \begin{subfigure}[b]{.5\columnwidth}
      \includegraphics[width=\textwidth]{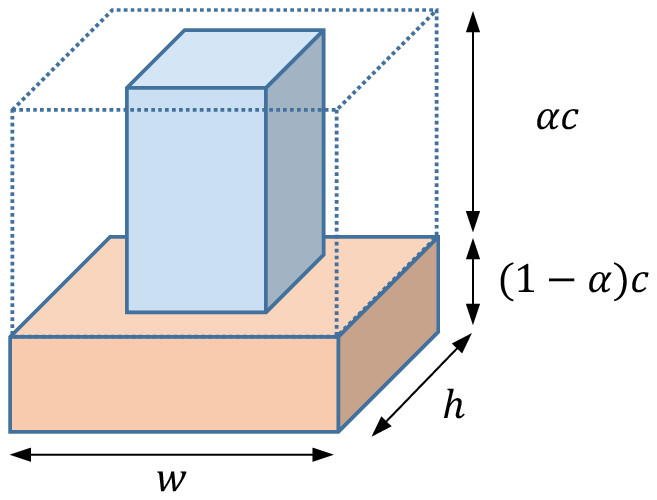}
      \caption{}
      \label{sup:fig:teaser_b}
    \end{subfigure}
    \caption{(a)~The conventional feature representation used by vanilla convolution. (c) The proposed multi-frequency feature representation stores the smoothly changing, low-frequency maps in a low-resolution tensor to reduce spatial redundancy, used by \hiConvName. The figure is rotated compared to the one in the main paper for clarity.}
    \label{sup:fig:teaser}
    
  \end{figure}

\myparagraph{Memory cost}
The proposed \hiConv stores the feature representation in a multi-frequency feature representation as shown in Figure~\ref{sup:fig:teaser}, where the low frequency tensor is stored in $2\times$ lower spatial resolution and thus cost $75\%$ less space to store the low frequency maps compared with the conventional feature representation. The relative memory cost is conditional on the ratio ($\alpha$) and is calculated by
\begin{equation}
    \begin{aligned}
    1 - \frac{3}{4}\alpha .
    \end{aligned}
    \label{eq:hiconv_mem-cost}
\end{equation}

\myparagraph{Computational cost}
The computational cost of \hiConv is proportional to the number of locations and channels that are needed to be convolved on. Following the design shown in Figure 2 in the main paper, we need to compute four paths, namely $H \rightarrow H$, $H \rightarrow L$, $L \rightarrow H$, and $L \rightarrow L$.

We assume the convolution kernel size is $k\times k$, the spatial resolution of the high-frequency feature is $h \times w$, and there are $(1-\alpha)c$  channels in the high-frequency part  and $\alpha c$ channels in the low-frequency part. Then the FLOPS for computing each paths are calculated as below.
\begin{equation}
\label{eqn: flops individual map}
    \begin{aligned}
    & FLOPS(Y^{H \rightarrow H}) = h \times w \times k^2 \times (1-\alpha)^2 \times c^2 \\
    & FLOPS(Y^{H \rightarrow L}) = \frac{h}{2} \times \frac{w}{2} \times k^2 \times \alpha \times (1-\alpha) \times c^2 \\
    & FLOPS(Y^{L \rightarrow H}) = \frac{h}{2} \times \frac{w}{2} \times k^2\times (1-\alpha) \times \alpha \times c^2 \\
    & FLOPS(Y^{L \rightarrow L}) = \frac{h}{2} \times \frac{w}{2} \times k^2 \times \alpha^2 \times c^2
    \end{aligned}
\end{equation}

We omit FLOPS for adding $Y^{H \rightarrow H}$ and $Y^{L \rightarrow H}$ together, as well as that of adding $Y^{L \rightarrow L}$ and $Y^{H \rightarrow H}$ together, since the FLOPS of such addition is less than $h \times w \times c$, and is negligible compared with other computational costs. The computational cost of the pooling operation is also ignorable compared with other computational cost. The nearest neighborhood up-sampling is basically duplicating values which does not involves any computational cost. Therefore, by adding up all FLOPS in Eqn~\ref{eqn: flops individual map}, we can estimate the overall FLOPS for compute $Y^H$ and $Y^L$ in Eqn~\ref{eqn: overall oct conv flops}. 

\begin{equation}
    \begin{aligned}
    & FLOPS([Y^H, Y^L]) = (1 - \frac{3}{4}\alpha(2-\alpha)) \times h \times w \times k^2 \times c^2
    \end{aligned}
\label{eqn: overall oct conv flops}
\end{equation}

For vanilla convolution, the FLOPS for computing output feature map $Y$ of size $c \times h \times w$ with the kernel size $k \times k$, and input feature map of size $c \times h \times w$, can be estimated as below.

\begin{equation}
    \begin{aligned}
    & FLOPS(Y) = h \times w \times k^2 \times  c^2
    \end{aligned}
\end{equation}

three out of four internal convolution operations are conducted on the lower resolution tensors except the first convolution, \emph{i.e.} $f(X^H, W^{H\xrightarrow{}H})$. Thus, the relative computational cost compared with vanilla convolution using the same kernel size and number of input/out channels is:
Therefore, the computational cost ratio between the \hiConv and vanilla convolution is $(1 - \frac{3}{4}\alpha(2-\alpha))$.

\begin{equation}
    \begin{aligned}
    & \frac{(1-\alpha)^2c^2 + \frac{1}{2}\alpha(1-\alpha)c^2 + \frac{1}{4}\alpha^2c^2}{c^2} \\
    & = 1 - \frac{3}{4}\alpha(2-\alpha) .
    \end{aligned}
    \label{eq:hiconv_flops-cost}
\end{equation}

Note that the computational cost of the pooling operation is ignorable and thus is not considered. The nearest neighborhood up-sampling is basically duplicating values which does not involves any computational cost.

%% file: supp/t2_imnet_results.tex
\section*{Appendix C. ImageNet Ablation Study Results}

Table~\ref{tab:cls:receptive-field} shows that the gain of OctConv over baseline models increases as the test image resolution grows. Such ability of better detecting large objects can be explained as the larger receptive field of each OctConv.

Table~\ref{tab:cls:ablation:center-shifting} shows an ablation study to examine down-sampling and inter-octave connectivity on ImageNet. The results confirm the importance of having both inter-frequency communication paths. It also shows that pooling methods are better than strided convolution and the average pooling works the best.

Table~\ref{tab:fig4} reports the values that are plotted in Figure 4 of the main text for clarity of presentation and to allow future work to compare to the precise numbers.

\begin{table}[h]
\centering
\renewcommand{\arraystretch}{1.2}
\resizebox{\columnwidth}{!}{
  \begin{tabular}{lccccccccc}
   \toprule
    \multirow{2}{*}{~Model~} & \multirow{2}{*}{~ratio ($\alpha$)~} & \multicolumn{7}{c}{ Testing Scale (\textit{small} $\xrightarrow{}$ \textit{large}) } \\
    \cline{3-10}
                     &      & ~$256$~ & ~$320$~ & ~$384$~ & ~$448$~ & ~$512$~ & ~$576$~ & ~$640$~ & ~$740$~ \\
   \midrule
      ResNet-50      &  N/A &   77.2  &  78.6   &  78.7   &   78.7  &   78.3  &   77.6  &   76.7  &   75.8  \\
   \hiConvPrefix-ResNet-50  & .5  & \tbf{+0.7} & \tbf{+0.7} & \tbf{+0.9} & \tbf{+0.9} & \tbf{+0.8} & \tbf{+1.0} & \tbf{+1.1} & \tbf{+1.2} \\
  \bottomrule
  \end{tabular}
}
\caption{ImageNet classification accuracy. The short length of input images are resized to the target crop size while keeping the aspect ratio unchanged. A centre crop is adopted if the input image size is not square. ResNet-50 backbone trained with crops size of $256 \times 256$ pixels.}
\label{tab:cls:receptive-field}
\end{table}

\begin{table}[h]
\centering
\renewcommand{\arraystretch}{1.2}
\resizebox{\columnwidth}{!}{
  \begin{tabular}{lcccc}
   \toprule
   Method          &    Down-sampling    &  Low $\xrightarrow{}$ High  &  High $\xrightarrow{}$ Low  &  Top-1 (\%) \\
   \midrule
   \multirow{5}{*}{\makecell[l]{\hiConvPrefix-ResNet-50\\ ratio: 0.5}}
                   &  avg. pooling    &                     &                      &  76.0   \\
                   &  avg. pooling    &     \checkmark      &                      &  76.4   \\
                   &  avg. pooling    &                     &      \checkmark      &  76.4   \\
                   \cmidrule{2-5}
                   & strided conv.    &     \checkmark      &      \checkmark      &  76.3   \\
                   &  max. pooling    &     \checkmark      &      \checkmark      &  77.0   \\
                   &  avg. pooling    &     \checkmark      &      \checkmark      &  77.4   \\
  \bottomrule
  \end{tabular}
}
\caption{Ablation on down-sampling and inter-octave connectivity on ImageNet. Note that MG-Conv~\cite{ke2017multigrid} uses max pooling for down-sampling.}
\label{tab:cls:ablation:center-shifting}
\end{table}

\begin{table}[t]
    \centering
    \renewcommand{\arraystretch}{1.2}
    \resizebox{\columnwidth}{!}{
    \begin{tabular}{lcccccc}
        \toprule
         Backbone & & baseline & $\alpha=0.125$ & $\alpha=0.25$ & $\alpha=0.5$ & $\alpha=0.75$ \\ \midrule
        \multirow{2}{*}{ResNet-26} &    GFLOPs &    
        2.353  & 2.102 & 1.871 & 1.491 & 1.216  \\ 
        & Top-1 acc. & 
        73.2   & 75.8  & 76.1  & 75.5  & 74.6 	\\ \midrule

        \multirow{2}{*}{DenseNet-121} &    GFLOPs &    
        2.852  & 2.428 & 2.044 &  -    &   -    \\
        & Top-1 acc. & 
        75.4   & 76.1  & 75.9  &  -    &   -
    	\\ \midrule   
    	
        \multirow{2}{*}{ResNet-50} &    GFLOPs &    
        4.105  & 3.587 & 3.123 & 2.383 & 1.891   \\ 
        & Top-1 acc. & 
        77.0   & 78.2  & 78.0  & 77.4  & 76.7	
    	\\ \midrule        

        \multirow{2}{*}{SE-ResNet-50} &    GFLOPs &    
        4.113  & 3.594 &  3.130	&  2.389 &  1.896	\\ 
        & Top-1 acc. & 
        77.6   & 78.7  &  78.4  &  77.9  & 77.4
    	\\ \midrule 
        
        \multirow{2}{*}{ResNeXt-50} &    GFLOPs &    
        4.250  & - & 3.196  & 2.406  & 1.891  	\\ 
        & Top-1 acc. & 
         78.4 & - & 78.8 & 78.4 & 77.5
    	\\ \midrule        

        \multirow{2}{*}{ResNet-101} &    GFLOPs &    
        7.822   & 6.656  & 5.625  & 4.012  & - \\ 
        & Top-1 acc. & 
        78.5  & 79.2 & 79.2   & 78.7  & -
    	\\ \midrule   
    	
        \multirow{2}{*}{ResNeXt-101} &    GFLOPs &    
        7.993  &  - &  5.719  &   4.050  & -  \\
        & Top-1 acc. & 
        79.4  &  - & 79.6  & 78.9 & - 
    	\\ \midrule        

        \multirow{2}{*}{ResNet-200} &    GFLOPs &    
        15.044  & 12.623  & 10.497  & 7.183   & - \\
        & Top-1 acc. & 
        79.6 & 80.0 & 79.8 & 79.5  & -
    	\\
        
        \bottomrule
    \end{tabular}
    }
    \caption{Ablation study on ImageNet in table form corresponding to the plots in Figure 4 in the main paper. Note: All networks are trained with na\"ive softmax loss without label smoothing~\cite{InceptionV4} or mixup~\cite{mixup}}
    \label{tab:fig4}
\end{table}

%% file: main.bbl
\begin{thebibliography}{10}\itemsep=-1pt

\bibitem{campbell1968application}
Fergus~W Campbell and JG Robson.
\newblock Application of fourier analysis to the visibility of gratings.
\newblock {\em The Journal of physiology}, 197(3):551--566, 1968.

\bibitem{k600}
Joao Carreira, Eric Noland, Andras Banki-Horvath, Chloe Hillier, and Andrew
  Zisserman.
\newblock A short note about kinetics-600.
\newblock {\em arXiv preprint arXiv:1808.01340}, 2018.

\bibitem{k400}
Joao Carreira and Andrew Zisserman.
\newblock Quo vadis, action recognition? a new model and the kinetics dataset.
\newblock In {\em proceedings of the IEEE Conference on Computer Vision and
  Pattern Recognition}, pages 6299--6308, 2017.

\bibitem{bLNet}
Chun-Fu Chen, Quanfu Fan, Neil Mallinar, Tom Sercu, and Rogerio Feris.
\newblock Big-little net: An efficient multi-scale feature representation for
  visual and speech recognition.
\newblock {\em Proceedings of the Seventh International Conference on Learning
  Representations}, 2019.

\bibitem{tvm}
Tianqi Chen, Thierry Moreau, Ziheng Jiang, Lianmin Zheng, Eddie Yan, Haichen
  Shen, Meghan Cowan, Leyuan Wang, Yuwei Hu, Luis Ceze, et~al.
\newblock $\{$TVM$\}$: An automated end-to-end optimizing compiler for deep
  learning.
\newblock In {\em 13th $\{$USENIX$\}$ Symposium on Operating Systems Design and
  Implementation ($\{$OSDI$\}$ 18)}, pages 578--594, 2018.

\bibitem{multifiber}
Yunpeng Chen, Yannis Kalantidis, Jianshu Li, Shuicheng Yan, and Jiashi Feng.
\newblock Multi-fiber networks for video recognition.
\newblock In {\em Proceedings of the European Conference on Computer Vision
  (ECCV)}, pages 352--367, 2018.

\bibitem{DPN}
Yunpeng Chen, Jianan Li, Huaxin Xiao, Xiaojie Jin, Shuicheng Yan, and Jiashi
  Feng.
\newblock Dual path networks.
\newblock In {\em Advances in Neural Information Processing Systems}, pages
  4467--4475, 2017.

\bibitem{glorea}
Yunpeng Chen, Marcus Rohrbach, Zhicheng Yan, Shuicheng Yan, Jiashi Feng, and
  Yannis Kalantidis.
\newblock Graph-based global reasoning networks.
\newblock In {\em Proceedings of the IEEE Conference on Computer Vision and
  Pattern Recognition}, 2019.

\bibitem{xception}
Fran{\c{c}}ois Chollet.
\newblock Xception: Deep learning with depthwise separable convolutions.
\newblock In {\em Proceedings of the IEEE conference on computer vision and
  pattern recognition}, pages 1251--1258, 2017.

\bibitem{spatialvision}
Russell~L. De~Valois and Karen~K. De~Valois.
\newblock Spatial vision.
\newblock {\em Oxford psychology series, No. 14.}, 1988.

\bibitem{imagenet}
Jia Deng, Wei Dong, Richard Socher, Li-Jia Li, Kai Li, and Li Fei-Fei.
\newblock Imagenet: A large-scale hierarchical image database.
\newblock In {\em 2009 IEEE conference on computer vision and pattern
  recognition}, pages 248--255. Ieee, 2009.

\bibitem{SlowFast}
Christoph Feichtenhofer, Haoqi Fan, Jitendra Malik, and Kaiming He.
\newblock Slowfast networks for video recognition.
\newblock {\em ICCV}, 2019.

\bibitem{imnet1hour}
Priya Goyal, Piotr Doll{\'a}r, Ross Girshick, Pieter Noordhuis, Lukasz
  Wesolowski, Aapo Kyrola, Andrew Tulloch, Yangqing Jia, and Kaiming He.
\newblock Accurate, large minibatch sgd: Training imagenet in 1 hour.
\newblock {\em arXiv preprint arXiv:1706.02677}, 2017.

\bibitem{gluoncvnlp2019}
Jian Guo, He He, Tong He, Leonard Lausen, Mu Li, Haibin Lin, Xingjian Shi,
  Chenguang Wang, Junyuan Xie, Sheng Zha, Aston Zhang, Hang Zhang, Zhi Zhang,
  Zhongyue Zhang, and Shuai Zheng.
\newblock Gluoncv and gluonnlp: Deep learning in computer vision and natural
  language processing.
\newblock {\em arXiv preprint arXiv:1907.04433}, 2019.

\bibitem{DSD}
Song Han, Jeff Pool, Sharan Narang, Huizi Mao, Enhao Gong, Shijian Tang, Erich
  Elsen, Peter Vajda, Manohar Paluri, John Tran, et~al.
\newblock Dsd: Dense-sparse-dense training for deep neural networks.
\newblock {\em arXiv preprint arXiv:1607.04381}, 2016.

\bibitem{ResNetV1}
Kaiming He, Xiangyu Zhang, Shaoqing Ren, and Jian Sun.
\newblock Deep residual learning for image recognition.
\newblock In {\em Proceedings of the IEEE Conference on Computer Vision and
  Pattern Recognition}, pages 770--778, 2016.

\bibitem{ResNetV2}
Kaiming He, Xiangyu Zhang, Shaoqing Ren, and Jian Sun.
\newblock Identity mappings in deep residual networks.
\newblock In {\em European conference on computer vision}, pages 630--645.
  Springer, 2016.

\bibitem{MobileNetV1}
Andrew~G Howard, Menglong Zhu, Bo Chen, Dmitry Kalenichenko, Weijun Wang,
  Tobias Weyand, Marco Andreetto, and Hartwig Adam.
\newblock Mobilenets: Efficient convolutional neural networks for mobile vision
  applications.
\newblock {\em arXiv preprint arXiv:1704.04861}, 2017.

\bibitem{SENet}
Jie Hu, Li Shen, and Gang Sun.
\newblock Squeeze-and-excitation networks.
\newblock In {\em Proceedings of the IEEE conference on computer vision and
  pattern recognition}, pages 7132--7141, 2018.

\bibitem{huang2018multi}
Gao Huang, Danlu Chen, Tianhong Li, Felix Wu, Laurens van~der Maaten, and
  Kilian~Q Weinberger.
\newblock Multi-scale dense networks for resource efficient image
  classification.
\newblock {\em ICLR}, 2018.

\bibitem{CondenseNet}
Gao Huang, Shichen Liu, Laurens Van~der Maaten, and Kilian~Q Weinberger.
\newblock Condensenet: An efficient densenet using learned group convolutions.
\newblock In {\em Proceedings of the IEEE Conference on Computer Vision and
  Pattern Recognition}, pages 2752--2761, 2018.

\bibitem{densenet}
Gao Huang, Zhuang Liu, Laurens Van Der~Maaten, and Kilian~Q Weinberger.
\newblock Densely connected convolutional networks.
\newblock In {\em Proceedings of the IEEE conference on computer vision and
  pattern recognition}, pages 4700--4708, 2017.

\bibitem{mkldnn}
Intel.
\newblock Math kernel library for deep neural networks (mkldnn).
\newblock
  https://github.com/intel/mkl-dnn/tree/7de7e5d02bf687f971e7668963649728356e0c20,
  2018.

\bibitem{kay2017kinetics}
Will Kay, Joao Carreira, Karen Simonyan, Brian Zhang, Chloe Hillier, Sudheendra
  Vijayanarasimhan, Fabio Viola, Tim Green, Trevor Back, Paul Natsev, et~al.
\newblock The kinetics human action video dataset.
\newblock {\em arXiv preprint arXiv:1705.06950}, 2017.

\bibitem{ke2017multigrid}
Tsung-Wei Ke, Michael Maire, and Stella~X Yu.
\newblock Multigrid neural architectures.
\newblock In {\em Proceedings of the IEEE Conference on Computer Vision and
  Pattern Recognition}, pages 6665--6673, 2017.

\bibitem{alexnet}
Alex Krizhevsky, Ilya Sutskever, and Geoffrey~E Hinton.
\newblock Imagenet classification with deep convolutional neural networks.
\newblock In {\em Advances in neural information processing systems}, pages
  1097--1105, 2012.

\bibitem{FPN}
Tsung-Yi Lin, Piotr Doll{\'a}r, Ross Girshick, Kaiming He, Bharath Hariharan,
  and Serge Belongie.
\newblock Feature pyramid networks for object detection.
\newblock In {\em Proceedings of the IEEE Conference on Computer Vision and
  Pattern Recognition}, pages 2117--2125, 2017.

\bibitem{lindeberg2013scale}
Tony Lindeberg.
\newblock {\em Scale-space theory in computer vision}, volume 256.
\newblock Springer Science \& Business Media, 2013.

\bibitem{PNASNet}
Chenxi Liu, Barret Zoph, Maxim Neumann, Jonathon Shlens, Wei Hua, Li-Jia Li, Li
  Fei-Fei, Alan Yuille, Jonathan Huang, and Kevin Murphy.
\newblock Progressive neural architecture search.
\newblock In {\em Proceedings of the European Conference on Computer Vision
  (ECCV)}, pages 19--34, 2018.

\bibitem{lowe2004distinctive}
David~G Lowe.
\newblock Distinctive image features from scale-invariant keypoints.
\newblock {\em International journal of computer vision}, 60(2):91--110, 2004.

\bibitem{ThiNet}
Jian-Hao Luo, Hao Zhang, Hong-Yu Zhou, Chen-Wei Xie, Jianxin Wu, and Weiyao
  Lin.
\newblock Thinet: pruning cnn filters for a thinner net.
\newblock {\em IEEE Transactions on Pattern Analysis and Machine Intelligence},
  2018.

\bibitem{ShuffleNetV2}
Ningning Ma, Xiangyu Zhang, Hai-Tao Zheng, and Jian Sun.
\newblock Shufflenet v2: Practical guidelines for efficient cnn architecture
  design.
\newblock In {\em Proceedings of the European Conference on Computer Vision
  (ECCV)}, pages 116--131, 2018.

\bibitem{AmoebaNet}
Esteban Real, Alok Aggarwal, Yanping Huang, and Quoc~V Le.
\newblock Regularized evolution for image classifier architecture search.
\newblock {\em Proceedings of the Thirty-Third AAAI Conference on Artificial
  Intelligence}, 2019.

\bibitem{MobileNetV2}
Mark Sandler, Andrew Howard, Menglong Zhu, Andrey Zhmoginov, and Liang-Chieh
  Chen.
\newblock Mobilenetv2: Inverted residuals and linear bottlenecks.
\newblock In {\em Proceedings of the IEEE Conference on Computer Vision and
  Pattern Recognition}, pages 4510--4520, 2018.

\bibitem{vgg}
Karen Simonyan and Andrew Zisserman.
\newblock Very deep convolutional networks for large-scale image recognition.
\newblock {\em arXiv preprint arXiv:1409.1556}, 2014.

\bibitem{HetConv}
Pravendra Singh, Vinay~Kumar Verma, Piyush Rai, and Vinay~P Namboodiri.
\newblock Hetconv: Heterogeneous kernel-based convolutions for deep cnns.
\newblock {\em arXiv preprint arXiv:1903.04120}, 2019.

\bibitem{stephane1999wavelet}
Mallat Stephane.
\newblock A wavelet tour of signal processing.

\bibitem{sun2019deep}
Ke Sun, Bin Xiao, Dong Liu, and Jingdong Wang.
\newblock Deep high-resolution representation learning for human pose
  estimation.
\newblock In {\em CVPR}, 2019.

\bibitem{sweldens1998lifting}
Wim Sweldens.
\newblock The lifting scheme: A construction of second generation wavelets.
\newblock {\em SIAM journal on mathematical analysis}, 29(2):511--546, 1998.

\bibitem{InceptionV4}
Christian Szegedy, Sergey Ioffe, Vincent Vanhoucke, and Alexander~A Alemi.
\newblock Inception-v4, inception-resnet and the impact of residual connections
  on learning.
\newblock In {\em Thirty-First AAAI Conference on Artificial Intelligence},
  2017.

\bibitem{InceptionV1}
Christian Szegedy, Wei Liu, Yangqing Jia, Pierre Sermanet, Scott Reed, Dragomir
  Anguelov, Dumitru Erhan, Vincent Vanhoucke, and Andrew Rabinovich.
\newblock Going deeper with convolutions.
\newblock In {\em Proceedings of the IEEE conference on computer vision and
  pattern recognition}, pages 1--9, 2015.

\bibitem{ClipQ}
Frederick Tung and Greg Mori.
\newblock Clip-q: Deep network compression learning by in-parallel
  pruning-quantization.
\newblock In {\em Proceedings of the IEEE Conference on Computer Vision and
  Pattern Recognition}, pages 7873--7882, 2018.

\bibitem{ELASTIC}
Huiyu Wang, Aniruddha Kembhavi, Ali Farhadi, Alan Yuille, and Mohammad
  Rastegari.
\newblock Elastic: Improving cnns with instance specific scaling policies.
\newblock {\em arXiv preprint arXiv:1812.05262}, 2018.

\bibitem{nonlocal}
Xiaolong Wang, Ross Girshick, Abhinav Gupta, and Kaiming He.
\newblock Non-local neural networks.
\newblock In {\em proceedings of the IEEE Conference on Computer Vision and
  Pattern Recognition}, 2017.

\bibitem{nonlocal-git}
Xiaolong Wang, Ross Girshick, Abhinav Gupta, and Kaiming He.
\newblock https://github.com/facebookresearch/video-nonlocal-net, 2018.

\bibitem{intensity}
Samuel~Webb Williams.
\newblock {\em Auto-tuning performance on multicore computers}.
\newblock University of California, Berkeley, 2008.

\bibitem{ResNeXt}
Saining Xie, Ross Girshick, Piotr Doll{\'a}r, Zhuowen Tu, and Kaiming He.
\newblock Aggregated residual transformations for deep neural networks.
\newblock In {\em Proceedings of the IEEE Conference on Computer Vision and
  Pattern Recognition}, pages 1492--1500, 2017.

\bibitem{mixup}
Hongyi Zhang, Moustapha Cisse, Yann~N Dauphin, and David Lopez-Paz.
\newblock mixup: Beyond empirical risk minimization.
\newblock {\em Proceedings of the Sixth International Conference on Learning
  Representations}, 2018.

\bibitem{ShuffleNetV1}
Xiangyu Zhang, Xinyu Zhou, Mengxiao Lin, and Jian Sun.
\newblock Shufflenet: An extremely efficient convolutional neural network for
  mobile devices.
\newblock In {\em Proceedings of the IEEE Conference on Computer Vision and
  Pattern Recognition}, pages 6848--6856, 2018.

\bibitem{PSP}
Hengshuang Zhao, Jianping Shi, Xiaojuan Qi, Xiaogang Wang, and Jiaya Jia.
\newblock Pyramid scene parsing network.
\newblock In {\em Proceedings of the IEEE conference on computer vision and
  pattern recognition}, pages 2881--2890, 2017.

\bibitem{NASNet}
Barret Zoph, Vijay Vasudevan, Jonathon Shlens, and Quoc~V Le.
\newblock Learning transferable architectures for scalable image recognition.
\newblock In {\em Proceedings of the IEEE conference on computer vision and
  pattern recognition}, pages 8697--8710, 2018.

\end{thebibliography}
